\documentclass[11pt]{article}
\usepackage[preprint]{acl}

\usepackage{times}
\usepackage[T1]{fontenc}
\usepackage[utf8]{inputenc}
\usepackage{microtype}
\usepackage{amsmath}
\usepackage{amssymb}
\usepackage{graphicx}
\usepackage{booktabs}
\usepackage{array}
\usepackage{enumitem}
\usepackage{xcolor}
\usepackage{multirow}
\usepackage{hyperref}
\usepackage{newunicodechar}
\newunicodechar{–}{--}

\usepackage{amsmath,amsfonts,bm}

\def\eqref#1{equation~\ref{#1}}
\def\1{\bm{1}}

\DeclareMathAlphabet{\mathsfit}{\encodingdefault}{\sfdefault}{m}{sl}
\SetMathAlphabet{\mathsfit}{bold}{\encodingdefault}{\sfdefault}{bx}{n}

\graphicspath{{figures/}}
\newcolumntype{L}[1]{>{\raggedright\arraybackslash}p{#1}}

\makeatletter\let\inputrows\@@input\makeatother % primitive \input: usable inside tabular
\newcommand{\sd}[1]{{\scriptsize$\pm$#1}}
\definecolor{tkbg}{HTML}{F1F5FB}
\definecolor{tkframe}{HTML}{2A78D6}
\newcommand{\takeaway}[2]{\par\smallskip\noindent\fcolorbox{tkframe}{tkbg}{\parbox{\dimexpr\linewidth-2\fboxsep-2\fboxrule\relax}{\textbf{Takeaway #1.} #2}}\par\smallskip}
\newcommand{\pend}[1]{\textcolor{red}{#1}}

\newif\ifzshiedit
\zshiedittrue
\ifzshiedit
    \newcommand{\zs}[1]{{\color{cyan}{[Zhouxing: #1]}}}
    
    \newcommand{\concern}[2]{\textcolor{red}{#1}{\textcolor{blue}{\ (ZS: #2)\ }}}
    
    \newcommand{\zsdel}[1]{\textcolor{gray}{[Delete] #1}}
\else
    \newcommand{\zs}[1]{}
    \newcommand{\concern}[2]{}
    
    \newcommand{\zsdel}[1]{}
\fi

\title{Where and What to Reward? Understanding Generalization Between Verifiable and Open-Ended Domains in Reinforcement Learning}

\author{
  \textbf{Shangjian Yin\textsuperscript{1}},
  \textbf{Yu Fu\textsuperscript{1}},
  \textbf{Yue Dong\textsuperscript{1}},
  \textbf{Zhouxing Shi\textsuperscript{1}}
\\
  \textsuperscript{1}University of California, Riverside
}

\begin{document}
\maketitle
% !TEX root = acl.tex
% v2 draft (analysis framing). The previous GRLO version is kept in main_grlo_v1.tex.
% Numbers come from scripts/v2/paper_numbers.py (which parses
% ~/verl_runs/final_summary_all_v2_20260929.md), scripts/v2/answer_only_by_epoch.json,
% or the files cited in the appendix. Run names map to the internal names listed in
% scripts/v2/make_figures.py (e.g., Chat-RM = "Ultra" in the result files).

% Keep the manuscript compatible with entry points predating the v2 helpers.
% Preserve any definitions already supplied by the entry point.
\providecolor{tkbg}{HTML}{F1F5FB}
\providecolor{tkframe}{HTML}{2A78D6}
\providecommand{\takeaway}[2]{\par\smallskip\noindent\fcolorbox{tkframe}{tkbg}{\parbox{\dimexpr\linewidth-2\fboxsep-2\fboxrule\relax}{\textbf{Takeaway #1.} #2}}\par\smallskip}
\providecommand{\sd}[1]{{\scriptsize$\pm$#1}}
\providecommand{\pend}[1]{\textcolor{red}{#1}}
\makeatletter
\@ifundefined{inputrows}{\let\inputrows\@@input}{}
\makeatother

\begin{abstract}
Reinforcement learning with verifiable rewards (RLVR) is widely used to improve the reasoning capability of language models, since model-generated answers can be easily verified by a rule-based verifier. In contrast, reinforcement learning from human feedback (RLHF) is mainly used to optimize open-ended conversation with a learned reward model, because such responses are not easy to verify. RLVR and RLHF are usually trained separately with their own domain data and training protocols. How gains transfer between these domains is unclear, partly because comparisons change both where a model is trained (the prompts) and what is rewarded (by a verifier or a reward model). We separate the two in a controlled study of RL on multiple base models that fixes the optimizer and the number of training prompts, and varies only the source domain of the prompts (math, multi-domain, chat, or mixtures) and the reward. We find that RLHF on open-ended prompts substantially improves conversational ability and also brings large gains in reasoning and coding, whereas RLVR improves reasoning but its gains in open-ended conversation plateau far below those of RLHF, suggesting that RL on open-ended prompts with a reward model can serve as an effective approach for generalizing across domains.
\end{abstract}

\section{Introduction}
\label{sec:intro}

Reinforcement learning (RL) has become a central part of post-training for large language models (LLMs), and current practice is split between two families. RL with verifiable rewards (RLVR) trains on prompts whose model-generated answers can be checked by a simple rule-based verifier, most prominently in mathematics, and has driven much of the recent progress in reasoning~\citep{shao2024deepseekmathpushinglimitsmathematical,guo2025deepseek,xie2025logic,ma2025generalreasoneradvancingllmreasoning}. RL from human feedback (RLHF) is nowadays mainly used for open-ended conversation, whose responses are non-verifiable: they have no reference answer, and their quality depends on human preference, which cannot be checked by rules and is costly to collect at scale. RLHF therefore trains a reward model once on a large human preference dataset and uses it to score each response during RL~\citep{ouyang2022traininglanguagemodelsfollow,liu2025skywork}. Although production pipelines often combine the two~\citep{guo2025deepseek,yang2025qwen3technicalreport}, existing controlled studies typically investigate them separately, each with its own domain data, reward, and training protocol, and evaluate them mainly within their own domain. As a result, it remains unclear how their gains transfer across domains: does RLVR on mathematics also make a model a better conversational assistant, and does RLHF on open-ended conversation also make it better at reasoning?

One response to this domain-coverage gap is to extend RLVR beyond mathematics. General-Reasoner~\citep{ma2025generalreasoneradvancingllmreasoning} converts knowledge from many domains into verifiable question--answer pairs, enabling the model to learn multi-domain knowledge in an RLVR-style format. However, its training budget is very large: 230K prompts and about 1.5K GPU-hours for a 4B model. Moreover, much of open-ended conversation has no ground-truth response at all and is more naturally scored by a reward model~\citep{ouyang2022traininglanguagemodelsfollow,bhaskar2025languagemodelsthinkchat}. Whether broader verifiable coverage or reward-model training on open-ended prompts transfers better therefore remains an open question.

\begin{figure*}[t]
    \centering
    \includegraphics[width=\linewidth]{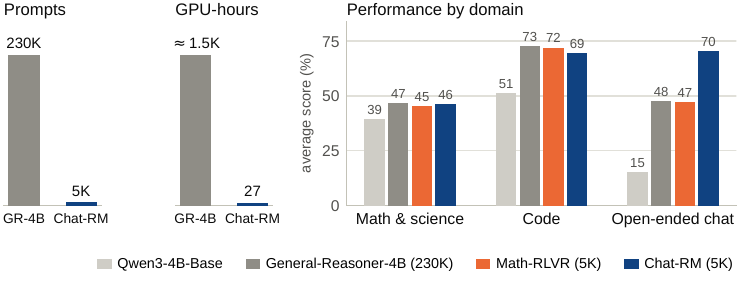}
    \caption{Training budget (left) and performance across domains (right) of General-Reasoner-4B and RL on Qwen3-4B-Base with 5K prompts (Math-RLVR and Chat-RM). Details are in \S\ref{sec:setup}.}
    \label{fig:overview}
\end{figure*}

This question is hard to answer from existing work, because RLVR and RLHF differ in two factors at once: \emph{where} the model is trained (the prompts) and \emph{what} is rewarded (by a verifier or a reward model). Comparing an RLVR model with an RLHF model therefore cannot tell which factor drives the transfer. In this work, we disentangle the two. We train multiple base models with RL under a single recipe, vary only the prompt source (math, multi-domain questions from General-Reasoner, chat prompts, or their mixtures) and the reward, and evaluate every checkpoint on both verifiable and open-ended benchmarks. Figure~\ref{fig:overview} previews the answer on Qwen3-4B-Base~\citep{yang2025qwen3technicalreport}: RLHF on chat prompts, which never sees a reference answer, performs on par with RLVR on math prompts in math and science (46 vs.\ 45 on average) and in code (69 vs.\ 72), while improving open-ended conversation far more (70 vs.\ 47). Both come close to General-Reasoner-4B on math and code with 46$\times$ fewer prompts and a small fraction of its compute. These results point to an asymmetry between the two domains, which we summarize as four takeaways, each detailed in its section:

\begin{enumerate}[leftmargin=*,itemsep=1pt,topsep=3pt,parsep=0pt,label=\textbf{T\arabic*.}]
\item For verifiable domains, neither \emph{where} nor \emph{what} (the reward type, verifier vs.\ reward model) matters much: chat prompts with a reward model nearly match math RLVR on math and code (\S\ref{sec:disentangle}).
\item For open-ended conversation, both matter: gains require a reward model and grow with how open-ended the prompts are, whereas with a verifier, open-ended quality stays low regardless of the prompts (\S\ref{sec:disentangle}). The asymmetry in T1 and T2 holds across the model sizes and reward models we study (\S\ref{sec:reward}, \S\ref{sec:8b}).
\item As a result, adding chat prompts to verifiable prompts under one reward model keeps the math gains and adds most of the open-ended gains (\S\ref{sec:combine}).
\item Open-ended gains keep growing with more chat prompts (and proportionally more training steps), whereas verifiable gains are largely in place with 1K prompts (\S\ref{sec:scale}).
\end{enumerate}

\section{Experimental Setup}
\label{sec:setup}

\paragraph{RLVR and RLHF.}
Both optimize a policy $\pi_\theta$ on a prompt distribution $\mathcal{D}$; they differ in the reward. Let $x\sim\mathcal{D}$ denote a prompt and $y\sim\pi_\theta(\cdot\mid x)$ a response sampled from the policy.
RLHF scores the whole response with a reward model $r_\phi$ trained on human preferences and regularizes the policy toward a reference policy $\pi_{\mathrm{ref}}$~\citep{ouyang2022traininglanguagemodelsfollow}:
\begin{equation}\label{eq:rlhf}
\begin{aligned}
&\max_\theta\;\mathbb{E}_{x\sim\mathcal{D}}\Big[\mathbb{E}_{y\sim\pi_\theta(\cdot\mid x)}\big[r_\phi(x,y)\big]\\
&\quad-\beta\,\mathrm{KL}\big(\pi_\theta(\cdot\mid x)\,\|\,\pi_{\mathrm{ref}}(\cdot\mid x)\big)\Big].
\end{aligned}
\end{equation}
It needs no reference answers and applies to any domain the reward model covers, including open-ended conversation. RLVR instead checks the final answer against a reference answer $a(x)$ and maximizes the verifiable reward $r_{\mathrm{ver}}(x,y)=\mathbf{1}[\mathrm{ans}(y)\equiv a(x)]$ in place of $r_\phi$; it therefore requires prompts with reference answers and applies only where answers can be verified. Both objectives can be optimized with policy-gradient methods such as PPO~\citep{ppo} or GRPO~\citep{shao2024deepseekmathpushinglimitsmathematical}.

\paragraph{Training setting.}
We conduct experiments mainly on Qwen3-4B-Base and Qwen3-8B-Base (\S\ref{sec:8b}), and also apply the same recipe to Llama-3.2-3B-Instruct (Appendix~\ref{app:llama}). To separate the effect of the reward from that of the optimizer, every run optimizes its own reward with the same PPO objective,
\begin{equation}\label{eq:ppo}
\max_\theta\;\mathbb{E}_{x,y,t}\big[\min\big(\rho_t\hat{A}_t,\,\mathrm{clip}(\rho_t,1\pm\epsilon)\hat{A}_t\big)\big],
\end{equation}
where the expectation is over prompts $x\sim\mathcal{D}$, responses $y\sim\pi_{\theta_{\mathrm{old}}}(\cdot\mid x)$, and the tokens $t$ of $y$ (each token in a batch is weighted equally), $\rho_t=\pi_\theta(y_t\mid x,y_{<t})/\pi_{\theta_{\mathrm{old}}}(y_t\mid x,y_{<t})$, and $\hat{A}_t$ is the generalized advantage estimate from a learned value function, with the reward given at the last token. We use a value critic initialized from the base model, generalized advantage estimation with $\gamma=\lambda=1$, $\beta=0$, and no entropy bonus; runs on Web prompts allow longer prompts (Appendix~\ref{app:training}). Each iteration samples one response per prompt for a batch of 1,024 prompts (temperature 1.0, at most 3,072 tokens) and makes one pass of PPO updates over mini-batches of 256 (actor learning rate $10^{-6}$, critic learning rate $10^{-5}$, clip ratio $\epsilon=0.2$). Each run trains on 5K prompts for 15 epochs (60 iterations), which takes 2.6--3.4 hours on one node with eight NVIDIA RTX PRO 6000 Blackwell GPUs for the 4B model. We use verl~\citep{sheng2024hybridflow}, and Appendix~\ref{app:training} lists all hyperparameters.

\paragraph{Where: prompt pools.}
We draw 5K prompts from each of the following three pools, ranging from verifiable to open-ended. \textbf{Math} consists of competition problems from the MATH training split~\citep{hendrycksmath2021}. \textbf{Web} consists of multi-domain WebInstruct questions released with General-Reasoner~\citep{ma2025generalreasoneradvancingllmreasoning}, mostly from STEM fields, with smaller shares from business, economics, and history. Both pools come with reference answers, and their prompts ask the model to reason step by step and put the final answer in \texttt{\textbackslash boxed\{\}}. \textbf{Chat} consists of open-ended conversation prompts randomly selected from UltraFeedback~\citep{cui2024ultrafeedback}; Appendix~\ref{app:prompts} analyzes their domains. The three pools thus cover single-domain RLVR, multi-domain RLVR, and RLHF on open-ended prompts that are multi-domain.

\paragraph{What: rewards.}
We use three kinds of reward. A \textbf{rule verifier} (for Math) gives 1 if the last boxed answer is equivalent to the reference and 0 otherwise. A \textbf{model verifier} (for Web), the 1.5B General-Verifier of \citet{ma2025generalreasoneradvancingllmreasoning}, compares the extracted final answer with the reference and gives $+1$ if they match, 0 if not, and $-0.5$ if no final answer can be extracted. A \textbf{reward model} (RM) scores the whole response given the prompt and never sees a reference answer. We use Skywork-Reward-V2-Llama-3.1-8B~\citep{liu2025skywork} by default and ablate different reward models in \S\ref{sec:reward}. Runs are named by prompt pool and reward; for example, Math-RLVR uses the rule verifier on Math prompts, and Math-RM uses the reward model on the same prompts. Math+Chat-RM mixes Math and Chat prompts in equal halves and uses the reward model on both, whereas Math-RLVR+Chat-RM and Web-RLVR+Chat-RM score their verifiable half with the verifier and their Chat half with the reward model; in these two runs, we divide the reward-model scores by 20 to bring them closer to the scale of the verifier rewards.

\paragraph{Evaluation.}
We evaluate math and science on MATH500~\citep{hendrycksmath2021,lightman2023let}, OlympiadBench~\citep{he2024olympiadbench}, Minerva Math~\citep{lewkowycz2022solving}, AIME 2024--2026~\citep{aime24,aime25,aime26} (mean over the three exams), and GPQA-Diamond~\citep{rein2023gpqa}; code on HumanEval and MBPP~\citep{chen2021codexhumaneval,austin2021mbpp} with the chat-style protocol of \citet{cui2025processreinforcementimplicitrewards}; and open-ended conversation on the AlpacaEval~2 length-controlled win rate (AE2 LC;~\citealp{dubois2025lengthcontrolledalpacaevalsimpleway}) and the Arena-Hard v0.1 win rate against GPT-4-0314~\citep{li2025arenahard}. Both open-ended benchmarks are judged by Qwen2-72B-Instruct~\citep{yang2024qwen2}, which \citet{yin-etal-2026-aligning} found to give AlpacaEval judgments comparable to GPT-4 at a much lower cost; we also report a second judge, Gemini 3.8 Flash, in \S\ref{sec:judge}. AIME averages 32 samples per problem and AlpacaEval uses one sample, both at temperature 0.6; all other benchmarks use greedy decoding (details in Appendix~\ref{app:eval}). Baselines are Qwen3-4B-Base, General-Reasoner-4B, and two SFT models trained for three epochs on 5K examples: Math-SFT on OpenR1-Math~\citep{openr1} solutions and Chat-SFT on UltraChat~\citep{ding-etal-2023-enhancing} conversations.

\begin{table*}[t]
    \centering
    \small
    \setlength{\tabcolsep}{1.6pt}
    \renewcommand{\arraystretch}{1.1}
    \begin{tabular*}{\linewidth}{@{\extracolsep{\fill}}l ccccccccc@{}}
\toprule
& \multicolumn{7}{c}{Verifiable domains} & \multicolumn{2}{c}{Open-ended domains} \\
\cmidrule(lr){2-8}\cmidrule(l){9-10}
Model & MATH500 & OlympiadBench & Minerva & AIME & GPQA & HumanEval & MBPP & AE2 LC & Arena-Hard \\
\midrule
\multicolumn{10}{@{}l}{\emph{Reference models}} \\
Qwen3-4B-Base & 71.6 & 38.7 & 35.7 & 7.7 & 43.4 & 58.5 & 44.1 & 4.3 & 26.5 \\
General-Reasoner-4B & 77.0 & 43.9 & 46.7 & \textbf{16.1} & \textbf{50.0} & \textbf{85.4} & 59.9 & 30.3 & 65.0 \\
\midrule
\multicolumn{10}{@{}l}{\emph{SFT (5K examples)}} \\
Math-SFT (OpenR1-Math) & 65.8 & 29.3 & 29.0 & 14.8 & 10.1 & 62.2 & 49.6 & 5.2 & 25.5 \\
Chat-SFT (UltraChat) & 46.2 & 20.4 & 24.6 & 2.2 & 30.8 & 64.6 & 51.9 & 2.6 & 12.6 \\
\midrule
\multicolumn{10}{@{}l}{\emph{RL on verifiable prompts}} \\
Math-RLVR & 79.8 & 41.3 & 47.8 & 14.3 & 43.4 & 82.9 & 61.1 & 29.9 & 64.2 \\
Web-RLVR & 77.6 & 42.4 & 49.3 & 12.1 & 38.9 & 84.2 & \textbf{62.2} & 35.1 & 65.9 \\
Math-RM & 78.8 & 38.8 & 47.8 & 12.5 & 49.0 & 81.1 & 57.6 & 34.3 & 65.8 \\
Web-RM & 78.6 & 41.3 & \textbf{50.7} & 11.4 & 45.5 & 78.0 & 56.1 & 47.8 & 73.7 \\
\midrule
\multicolumn{10}{@{}l}{\emph{RL on mixed prompts (half verifiable, half chat)}} \\
Web-RLVR+Chat-RM & 77.6 & 41.8 & 49.3 & 11.4 & 47.5 & 81.1 & 61.9 & 43.6 & 72.9 \\
Math-RLVR+Chat-RM & \textbf{81.0} & 42.5 & 47.8 & 14.1 & 46.5 & 81.7 & 59.6 & 46.5 & 76.3 \\
Math+Chat-RM & 78.6 & \textbf{44.3} & 46.3 & 12.9 & 49.0 & 81.1 & 56.6 & 54.0 & 76.6 \\
\midrule
\multicolumn{10}{@{}l}{\emph{RL on chat prompts}} \\
Chat-RM & 79.2 & 43.6 & 48.2 & 11.7 & 49.0 & 82.3 & 56.4 & \textbf{61.2} & \textbf{79.4} \\
\bottomrule
\end{tabular*}

    \caption{Main results (\%). All SFT and RL models are trained from Qwen3-4B-Base on 5K examples or prompts. Bold marks the best score in each column.}
    \label{tab:main}
\end{table*}

\section{Main Results}
\label{sec:results}

\subsection{Evaluation on the Qwen3-4B Backbone}
\label{sec:asym}

Table~\ref{tab:main} reports eight RL configurations trained from Qwen3-4B-Base, grouped by their prompts: verifiable prompts with a verifier (Math-RLVR, Web-RLVR) or the reward model (Math-RM, Web-RM); mixtures of verifiable and chat prompts (Web-RLVR+Chat-RM, Math-RLVR+Chat-RM, Math+Chat-RM); and chat prompts alone (Chat-RM). We compare them with the base model, General-Reasoner-4B, and two SFT baselines trained on the same budget. Across these configurations, verifiable and open-ended gains behave very differently.

\paragraph{Verifiable gains barely depend on the prompts.}
Every RL configuration improves on the base model on all math and code benchmarks, and by similar margins: MATH500 rises from 71.6 to 77.6--81.0, Minerva from 35.7 to 46.3--50.7, AIME from 7.7 to 11.4--14.3, HumanEval from 58.5 to 78.0--84.2, and MBPP from 44.1 to 56.1--62.2. This transfer runs in both directions. Chat-RM, which never sees a reference answer, matches Math-RLVR on MATH500 (79.2 vs.\ 79.8) and exceeds it on OlympiadBench (43.6 vs.\ 41.3), trailing only on AIME (11.7 vs.\ 14.3); domain-matched RL thus still helps on the hardest competition problems, but only by a small margin. Conversely, although the Math and Web pools contain almost no programming prompts, Math-RLVR and Web-RLVR gain as much on HumanEval ($+24.4$ and $+25.7$) as Chat-RM ($+23.8$). Most of the verifiable gain therefore transfers across domains.

\begin{figure}[t]
    \centering
    \includegraphics[width=\linewidth]{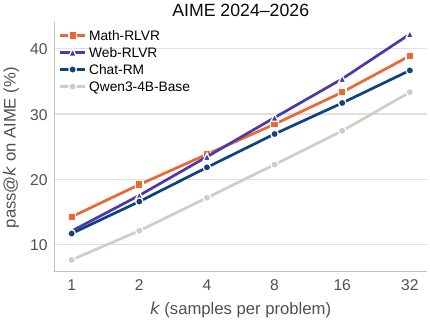}
    \caption{Unbiased pass@$k$ pooled over AIME 2024--2026 (90 problems, 32 samples per problem).}
    \label{fig:passk}
\end{figure}

\paragraph{The gains reflect better reasoning.}
An error analysis of MATH500 (Table~\ref{tab:errors_main}) attributes the gains to fewer reasoning errors rather than repaired answer formats: reasoning errors fall from the base model's 26.6\% to 19.6--21.6\% across the four RL runs (Math-RLVR, Web-RLVR, Math-RM, and Chat-RM), while correctly reasoned but mis-formatted answers stay at or below 0.8\% (1.8\% for the base model). Sampling-based metrics agree: pooled over AIME 2024--2026, Math-RLVR, Web-RLVR, and Chat-RM improve on the base model in pass@$k$ for every number of samples $k$ up to 32 (Figure~\ref{fig:passk}; Appendix~\ref{app:passk}). RL thus improves how the model reasons, not merely how it formats its answers.

\begin{table}[t]
    \centering
    \small
    \setlength{\tabcolsep}{3pt}
    \begin{tabular}{@{}lccc@{}}
\toprule
& & \multicolumn{2}{c}{Errors} \\
\cmidrule(l){3-4}
Model & Accuracy $\uparrow$ & Format $\downarrow$ & Reasoning $\downarrow$ \\
\midrule
Qwen3-4B-Base & 71.6 & 1.8 & 26.6 \\
\midrule
Math-RLVR & 79.8 & 0.6 & 19.6 \\
Web-RLVR & 77.6 & 0.8 & 21.6 \\
Chat-RM & 79.2 & 0.8 & 20.0 \\
Math-RM & 78.8 & 0.4 & 20.8 \\
\bottomrule
\end{tabular}

    \caption{MATH500 accuracy and error rates (\%). Format errors have correct reasoning but a mis-formatted answer (Appendix~\ref{app:errors}).}
    \label{tab:errors_main}
\end{table}

\begin{figure*}[t]
    \centering
    \includegraphics[width=\linewidth]{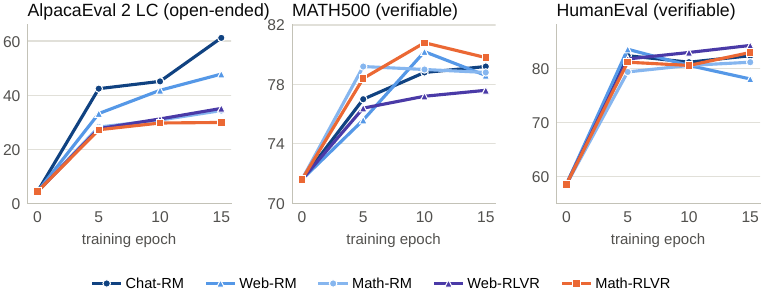}
    \caption{Training dynamics of five configurations (epoch 0 is the base model). They separate on AlpacaEval~2 LC from epoch 5 on but stay close on MATH500 and HumanEval (per-epoch scores in Appendix~\ref{app:epochs}).}
    \label{fig:dynamics}
\end{figure*}

\begin{figure}[t]
    \centering
    \includegraphics[width=\linewidth]{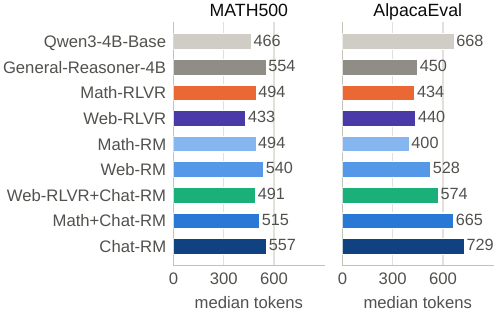}
    \caption{Median response length (tokens) on MATH500 and AlpacaEval.}
    \label{fig:length}
\end{figure}

\paragraph{Open-ended quality varies widely.}
In contrast, runs with nearly identical verifiable accuracy differ widely on open-ended benchmarks: AE2 LC ranges from 29.9 to 61.2 and Arena-Hard from 64.2 to 79.4. Reward-model runs do write longer AlpacaEval responses as the training prompts become more open-ended (Figure~\ref{fig:length}), but the gap is not an artifact of response length or style: it persists under Arena-Hard's style control, which discounts length and markdown formatting (77.0 for Chat-RM vs.\ 62.8 for Math-RLVR; Appendix~\ref{app:length}). It also emerges early in training (Figure~\ref{fig:dynamics}): the configurations separate on AlpacaEval from epoch 5 on, while MATH500 and HumanEval stay within a narrow band. Unlike verifiable ability, open-ended quality thus depends on how the model is trained, which we disentangle next.

\subsection{Disentangling Where and What}
\label{sec:disentangle}

The comparisons above vary both the prompts (\emph{where}) and the reward (\emph{what}) at once. We next disentangle the two by arranging the runs of Table~\ref{tab:main} by prompt source and reward.

\paragraph{Fixing the prompts, varying the reward.}
On Math prompts, replacing the rule verifier with the reward model changes little in either domain. Math-RM recovers most of the verifier's gain on MATH500 ($+7.2$ vs.\ $+8.2$ over the base model), Minerva ($+12.1$ for both), and AIME ($+4.8$ vs.\ $+6.6$), though not on OlympiadBench ($+0.1$ vs.\ $+2.6$), and adds only 4.4 points of AE2 LC and 1.6 points of Arena-Hard. On Web prompts, the same swap keeps every math score within 1.5 points but adds 12.7 points of AE2 LC and 7.8 points of Arena-Hard. Only code is consistently a few points lower with the reward model (MBPP 56.1--57.6 vs.\ 61.1--62.2). An explicit, task-specific correctness reward is therefore not needed for most of the verifiable gain. A possible reason is that a strong preference reward model may itself encode aspects of correctness and reasoning quality. The choice of reward thus matters for open-ended quality mainly when the prompts leave room for responses that differ beyond the final answer, as the multi-domain Web questions do more than competition math problems.

\takeaway{1}{Verifiable gains depend little on the prompt source or on the reward type (verifier vs.\ reward model): chat prompts with a reward model nearly match math RLVR on math and code.}

\paragraph{Fixing the reward, varying the prompts.}
With a verifier, open-ended quality stays low regardless of the prompts: Math-RLVR and Web-RLVR reach only 29.9 and 35.1 on AE2 LC and 64.2 and 65.9 on Arena-Hard, and General-Reasoner-4B, trained with the same model verifier on 46$\times$ more Web prompts, scores similarly (30.3 and 65.0). With the reward model, open-ended quality instead improves as the prompts become more open-ended, from Math to multi-domain Web prompts and then to 50\% and 100\% Chat prompts: AE2 LC is 34.3 for Math-RM, 47.8 for Web-RM, 54.0 for Math+Chat-RM, and 61.2 for Chat-RM, and Arena-Hard is 65.8, 73.7, 76.6, and 79.4, respectively. A verifier thus does not reward the qualities that open-ended evaluation measures, even with much more verifiable data, whereas a reward model improves them to the extent that its prompts cover open-ended requests.

\takeaway{2}{Open-ended gains require a reward model and increase with the share of open-ended prompts; with a verifier, open-ended quality stays low for all prompt sources.}

\subsection{Ablation on Different Reward Models}
\label{sec:reward}

Our main runs use Skywork-Reward-V2-Llama-3.1-8B as the reward model. To test whether the findings depend on this choice, Table~\ref{tab:reward} trains Chat-RM with two smaller reward models from the same Skywork-Reward-V2 family~\citep{liu2025skywork}, Skywork-Reward-V2-Llama-3.2-3B and Skywork-Reward-V2-Qwen3-4B, which are trained on the same preference data with different backbones, keeping everything else fixed. The smaller reward models match the 8B reward model on Arena-Hard (80.3 and 81.3 vs.\ 79.4) and, although lower on AE2 LC (54.0 and 53.5 vs.\ 61.2), remain far above the verifier runs (at most 35.1; Table~\ref{tab:main}). Their verifiable scores are similar (MATH500 78.6 and 80.0 vs.\ 79.2), with a slightly lower GPQA (41.9 and 42.9 vs.\ 49.0). A small reward model therefore suffices, and the asymmetry does not hinge on one particular reward model.

\begin{table*}[t]
    \centering
    \small
    \setlength{\tabcolsep}{1.6pt}
    \renewcommand{\arraystretch}{1.1}
    \begin{tabular*}{\linewidth}{@{\extracolsep{\fill}}l ccccccccc@{}}
\toprule
& \multicolumn{7}{c}{Verifiable domains} & \multicolumn{2}{c}{Open-ended domains} \\
\cmidrule(lr){2-8}\cmidrule(l){9-10}
Reward model & MATH500 & OlympiadBench & Minerva & AIME & GPQA & HumanEval & MBPP & AE2 LC & Arena-Hard \\
\midrule
None (Qwen3-4B-Base) & 71.6 & 38.7 & 35.7 & 7.7 & 43.4 & 58.5 & 44.1 & 4.3 & 26.5 \\
\midrule
3B (Llama-3.2) & 78.6 & \textbf{43.9} & \textbf{48.2} & \textbf{13.2} & 41.9 & 77.4 & 58.4 & 54.0 & 80.3 \\
4B (Qwen3) & \textbf{80.0} & 41.9 & 47.1 & 13.0 & 42.9 & 81.7 & \textbf{61.4} & 53.5 & \textbf{81.3} \\
8B (Llama-3.1) & 79.2 & 43.6 & \textbf{48.2} & 11.7 & \textbf{49.0} & \textbf{82.3} & 56.4 & \textbf{61.2} & 79.4 \\
\bottomrule
\end{tabular*}

    \caption{Reward-model ablation (\%): Chat-RM with Skywork-Reward-V2 reward models of different sizes. Bold marks the best score in each column.}
    \label{tab:reward}
\end{table*}

\begin{table*}[t]
    \centering
    \small
    \setlength{\tabcolsep}{1.6pt}
    \renewcommand{\arraystretch}{1.1}
    \begin{tabular*}{\linewidth}{@{\extracolsep{\fill}}l ccccccccc@{}}
\toprule
& \multicolumn{7}{c}{Verifiable domains} & \multicolumn{2}{c}{Open-ended domains} \\
\cmidrule(lr){2-8}\cmidrule(l){9-10}
Training & MATH500 & OlympiadBench & Minerva & AIME & GPQA & HumanEval & MBPP & AE2 LC & Arena-Hard \\
\midrule
\multicolumn{10}{@{}l}{\emph{One stage (Table~\ref{tab:main})}} \\
Math-RLVR & 79.8 & 41.3 & 47.8 & 14.3 & 43.4 & 82.9 & 61.1 & 29.9 & 64.2 \\
Web-RLVR & 77.6 & 42.4 & 49.3 & 12.1 & 38.9 & 84.2 & 62.2 & 35.1 & 65.9 \\
Chat-RM & 79.2 & \textbf{43.6} & 48.2 & 11.7 & \textbf{49.0} & 82.3 & 56.4 & \textbf{61.2} & 79.4 \\
\midrule
\multicolumn{10}{@{}l}{\emph{Verifier first, then reward model}} \\
Math-RLVR $\rightarrow$ Chat-RM & \textbf{83.2} & 43.1 & \textbf{51.5} & \textbf{14.7} & 44.4 & 84.8 & 59.9 & 48.8 & 80.3 \\
Web-RLVR $\rightarrow$ Chat-RM & 80.0 & 42.1 & 50.0 & 14.2 & 48.5 & 82.3 & \textbf{62.4} & 50.1 & \textbf{81.6} \\
\midrule
\multicolumn{10}{@{}l}{\emph{Reward model first, then verifier}} \\
Chat-RM $\rightarrow$ Math-RLVR & 80.4 & 43.0 & 48.2 & 13.2 & 41.9 & 80.5 & 57.4 & 52.2 & 80.2 \\
Chat-RM $\rightarrow$ Web-RLVR & 79.6 & 43.0 & 48.5 & 12.2 & 43.4 & \textbf{85.4} & 58.9 & 60.4 & 78.7 \\
\bottomrule
\end{tabular*}

    \caption{Sequential training (\%). The second stage starts from the final checkpoint of the first. Bold marks the best score in each column.}
    \label{tab:seq}
\end{table*}

\subsection{How About Joint Training?}
\begin{table}[t]
    \centering
    \small
    \setlength{\tabcolsep}{4pt}
    \begin{tabular}{@{}lcccc@{}}
        \toprule
        Prompts & MATH500 & GPQA & HumanEval & AE2 LC \\
        \midrule
        1K & 76.4 & \textbf{49.5} & \textbf{83.5} & 34.8 \\
        3K & 78.6 & 44.4 & 80.5 & 46.4 \\
        5K & \textbf{79.2} & 49.0 & 82.3 & \textbf{61.2} \\
        \bottomrule
    \end{tabular}
    \caption{Chat-RM with 1K, 3K, and 5K prompts (\%). Bold marks the best score in each column.}
    \label{tab:scale_main}
\end{table}

\begin{table*}[t]
    \centering
    \small
    \setlength{\tabcolsep}{1.6pt}
    \renewcommand{\arraystretch}{1.1}
    \begin{tabular*}{\linewidth}{@{\extracolsep{\fill}}l ccccccccc@{}}
\toprule
& \multicolumn{7}{c}{Verifiable domains} & \multicolumn{2}{c}{Open-ended domains} \\
\cmidrule(lr){2-8}\cmidrule(l){9-10}
Model & MATH500 & OlympiadBench & Minerva & AIME & GPQA & HumanEval & MBPP & AE2 LC & Arena-Hard \\
\midrule
Qwen3-8B-Base & 68.4 & 35.9 & 38.6 & 8.5 & 40.9 & 32.9 & 43.1 & 14.2 & 45.6 \\
\midrule
Web-RLVR & 79.2 & 44.0 & 51.5 & 12.9 & 44.9 & 84.8 & \textbf{66.9} & 32.5 & 66.9 \\
Math-RLVR & \textbf{82.4} & \textbf{46.5} & 49.6 & 13.5 & 43.4 & 87.2 & 65.7 & 35.6 & 69.0 \\
Math-RM & 79.8 & 43.6 & \textbf{51.8} & 14.9 & \textbf{47.5} & 88.4 & 66.4 & 39.1 & 74.7 \\
Chat-RM & 81.0 & 43.4 & 50.4 & \textbf{15.7} & 47.0 & \textbf{90.2} & 63.7 & \textbf{63.6} & \textbf{85.1} \\
\bottomrule
\end{tabular*}

    \caption{Results on Qwen3-8B-Base (\%), with the recipe and evaluation of Table~\ref{tab:main}. Bold marks the best score in each column.}
    \label{tab:8b}
\end{table*}

\label{sec:combine}

Chat-RM already obtains both kinds of gains, but practical pipelines also train on verifiable prompts, and a verifier still helps most on the hardest math problems (AIME 14.3 for Math-RLVR vs.\ 11.7 for Chat-RM; Table~\ref{tab:main}). We therefore study joint training on verifiable (Math or Web) and Chat prompts in three ways: mixing the prompts under one reward model, mixing them with a verifier for the verifiable half and the reward model for the Chat half, and applying the two rewards in sequence.

\paragraph{One reward model.}
Math+Chat-RM replaces half of the Math prompts with Chat prompts and scores both with the same reward model. Compared with Math-RM, it keeps the math scores (MATH500 78.6 vs.\ 78.8; AIME 12.9 vs.\ 12.5) and adds 19.7 points of AE2 LC and 10.8 points of Arena-Hard, recovering most of the open-ended quality of Chat-RM (Arena-Hard 76.6 vs.\ 79.4). Joint training under one reward model thus combines both gains without a trade-off in mathematics.

\paragraph{Mixed rewards.}
Math-RLVR+Chat-RM uses the same prompts as Math+Chat-RM but scores the Math half with the rule verifier instead of the reward model, so comparing the two isolates the effect of the reward on the Math half. The verifier raises MATH500 (81.0 vs.\ 78.6) and AIME (14.1 vs.\ 12.9) but lowers AE2 LC (46.5 vs.\ 54.0), leaving Arena-Hard unchanged (76.3 vs.\ 76.6). With Web prompts, mixed rewards are weaker on open-ended benchmarks (Web-RLVR+Chat-RM: AE2 LC 43.6, Arena-Hard 72.9). A verifier thus adds a little on competition math at some cost in AE2 LC.
% \zs{Table 1 is far away now (two pages away?). Consider putting a smaller table for this subsection around this part rather than putting every thing in Table 1, if these mixture experiments are not discussed earlier.}
% Shangjian: not enough page space for an extra table.

\paragraph{Sequential training.}
When the two rewards are applied in sequence, each stage starting from the final checkpoint of the previous one, the order decides which gain is kept (Table~\ref{tab:seq}). Verifier first preserves the math gains and adds most of the open-ended ones: Math-RLVR$\rightarrow$Chat-RM reaches the highest MATH500 (83.2) and AIME (14.7) among our 4B RL runs and an Arena-Hard of 80.3, but an AE2 LC of only 48.8. Reward model first preserves Arena-Hard (78.7--80.2) and MATH500 (79.6--80.4), but Math prompts lower AE2 LC from 61.2 to 52.2.

\takeaway{3}{Joint training obtains both: mixing chat prompts into verifiable RL under one reward model keeps the math gains and adds most open-ended gains.}

\subsection{Data Scale}
\label{sec:scale}

We next vary the number of Chat prompts, training Chat-RM on 1K, 3K, and 5K prompts for 15 epochs each, which amounts to 15, 30, and 60 iterations (Table~\ref{tab:scale_main}). Open-ended quality keeps improving and shows no sign of saturating: AE2 LC rises from 34.8 to 46.4 (+11.6) and then to 61.2 (+14.8). Already with 1K prompts, Chat-RM reaches the open-ended level of the verifier runs (34.8 vs.\ 29.9 for Math-RLVR and 35.1 for Web-RLVR), and with 3K it approaches Web-RM (46.4 vs.\ 47.8). Verifiable ability, in contrast, is largely in place after 1K prompts: MATH500 rises from 71.6 for the base model to 76.4 and changes little afterwards (78.6 and 79.2), and HumanEval at 1K (83.5) is already as high as at 5K (82.3). Because the number of epochs is fixed, larger pools also receive more iterations, so this comparison does not separate data from compute. It nevertheless shows that a small set of chat prompts already elicits most of the verifiable gain, whereas open-ended quality keeps benefiting from more of them.

\takeaway{4}{More chat prompts keep improving open-ended quality, while verifiable gains saturate early.}

\subsection{Evaluation on the Qwen3-8B Backbone}
\label{sec:8b}

To test whether the findings carry over to a larger model, we repeat four configurations on Qwen3-8B-Base with the same recipe and evaluation (Table~\ref{tab:8b}). The same asymmetry appears. On verifiable benchmarks, all four runs improve substantially over the base model and perform similarly, and Chat-RM again matches the verifier runs without seeing a reference answer. On open-ended benchmarks, the verifier runs stay far behind, while the reward-model runs improve as the prompts become more open-ended, with Chat-RM reaching the best scores. The open-ended gap between Chat-RM and the verifier runs is at least as large as for the 4B model, so the asymmetry is not specific to the 4B model.

\subsection{Evaluation with a Different Judge}
\label{sec:judge}

Our open-ended scores rely on Qwen2-72B-Instruct as the judge. To check that the conclusions do not depend on this judge, we re-judge three runs with Gemini 3.8 Flash (Table~\ref{tab:judge}). Gemini assigns lower absolute scores throughout but preserves the ordering: Chat-RM leads the verifier runs by 10.5--13.0 points on AE2 LC (26.1--31.3 with Qwen2) and by 7.6--7.7 points on Arena-Hard (13.5--15.2 with Qwen2), well outside Gemini's bootstrap intervals of about $\pm$2 points, while Math-RLVR and Web-RLVR remain tied on Arena-Hard. The open-ended advantage of reward-model RL is therefore not an artifact of the judge.

\begin{table}[t]
    \centering
    \small
    \setlength{\tabcolsep}{4pt}
    \begin{tabular}{@{}lcccc@{}}
        \toprule
        & \multicolumn{2}{c}{AE2 LC} & \multicolumn{2}{c}{Arena-Hard} \\
        \cmidrule(lr){2-3}\cmidrule(l){4-5}
        Model & Qwen2 & Gemini & Qwen2 & Gemini \\
        \midrule
        Math-RLVR & 29.9 & 8.9 & 64.2 & 32.8 \\
        Web-RLVR & 35.1 & 11.4 & 65.9 & 32.9 \\
        Chat-RM & \textbf{61.2} & \textbf{21.9} & \textbf{79.4} & \textbf{40.5} \\
        \bottomrule
    \end{tabular}
    \caption{Open-ended scores (\%) under two judges: Qwen2-72B-Instruct and Gemini 3.8 Flash. Bold marks the best score in each column.}
    \label{tab:judge}
\end{table}

\section{Related Work}
\label{sec:related}

\paragraph{RL for reasoning.}
RL against rule-based or verifier-based rewards has become the dominant way to strengthen mathematical and logical reasoning, from GRPO in DeepSeekMath~\citep{shao2024deepseekmathpushinglimitsmathematical} to RL applied directly to base models in DeepSeek-R1~\citep{guo2025deepseek}, together with work on credit assignment and reward construction~\citep{cheng2025pure,xie2025logic,lightman2023let,zhang2024generative,chen2025better}. General-Reasoner~\citep{ma2025generalreasoneradvancingllmreasoning} extends verifiable rewards beyond mathematics with model-verified questions from many domains. A parallel line distills long chains of thought through supervised fine-tuning~\citep{ye2025limo,muennighoff2025s1simpletesttimescaling,openr1,yeo2025demystifying,yu2025longshortchainofthoughtmixturesupervised}. These works mostly train and evaluate within verifiable domains.

\paragraph{Preference-based training.}
RL from human feedback optimizes a learned reward model on open-ended prompts~\citep{ouyang2022traininglanguagemodelsfollow}, and direct preference optimization and its variants remove the explicit reward model~\citep{rafailov2023direct,ethayarajh2024kto,azar2024general}. Iterative and self-improving variants have been applied to alignment and reasoning~\citep{singh2023beyond,xiong2024iterative,xiong2024building,chen2024self,chen2024bootstrapping,deng2024flow,liu2024iterative,tu2025enhancing}, and stronger open reward models and benchmarks for them have appeared~\citep{liu2025skywork,lambert2024rewardbench,cui2024ultrafeedback}. Closest to our chat runs, \citet{bhaskar2025languagemodelsthinkchat} train models to think before answering chat prompts with a reward model. Instead of proposing a recipe, we hold one recipe fixed and cross the prompt distribution with the reward type, measuring transfer in both directions.

\section{Conclusion}
\label{sec:conclusion}

We analyze how RL gains transfer between verifiable and open-ended domains by separating where a model is trained from what is rewarded under a single recipe. The transfer is asymmetric. Verifiable gains in math and code barely depend on either factor: RL on chat prompts with a reward model nearly matches math RLVR. Open-ended gains instead require a reward model and grow with how open-ended the prompts are, whereas a verifier plateaus regardless of the prompts. These findings hold across reward models, model sizes and families, and judges, and suggest that RL on open-ended prompts with a reward model is an efficient way to obtain gains in both domains.

\section*{Limitations}

We study what RL elicits from a base model in single-turn generation with a fixed generation budget in non-thinking mode. We do not explore additional reasoning budgets or thinking mode, nor extensions to long-horizon agentic tasks; we leave these to future work.

\section*{Ethical Considerations}

We use publicly released datasets and models under their licenses: MATH, UltraFeedback, and UltraChat (MIT); WebInstruct, OpenR1-Math, and the AIME sets (Apache~2.0); Qwen3-4B/8B-Base, General-Verifier, General-Reasoner-4B, and Skywork-Reward-V2-Qwen3-4B (Apache~2.0); Llama-3.2-3B-Instruct and the Llama-based Skywork-Reward-V2 models (Llama~3.1/3.2 Community License); and Qwen2-72B-Instruct (Tongyi Qianwen License). The study involves no human subjects. The reward models and judges we rely on encode the preferences of their training data, which can include social biases, and optimizing against them can amplify such biases; our checkpoints received no safety training and are research artifacts only. Each training run used about 20--27 GPU-hours.

\bibliography{colm2026_conference,example_paper}

\begin{thebibliography}{48}
\providecommand{\natexlab}[1]{#1}

\bibitem[{Austin et~al.(2021)Austin, Odena, Nye, Bosma, Michalewski, Dohan, Jiang, Cai, Terry, Le, and Sutton}]{austin2021mbpp}
Jacob Austin, Augustus Odena, Maxwell Nye, Maarten Bosma, Henryk Michalewski, David Dohan, Ellen Jiang, Carrie Cai, Michael Terry, Quoc Le, and Charles Sutton. 2021.
\newblock \href {https://arxiv.org/abs/2108.07732} {Program synthesis with large language models}.
\newblock \emph{Preprint}, arXiv:2108.07732.

\bibitem[{Bhaskar et~al.(2026)Bhaskar, Ye, and Chen}]{bhaskar2025languagemodelsthinkchat}
Adithya Bhaskar, Xi~Ye, and Danqi Chen. 2026.
\newblock Language models that think, chat better.
\newblock In \emph{Conference on Language Modeling (COLM)}.

\bibitem[{Chen et~al.(2025{\natexlab{a}})Chen, Liu, Du, Pang, Liu, Sinha, Varakantham, and Lin}]{chen2024bootstrapping}
Changyu Chen, Zichen Liu, Chao Du, Tianyu Pang, Qian Liu, Arunesh Sinha, Pradeep Varakantham, and Min Lin. 2025{\natexlab{a}}.
\newblock \href {https://proceedings.iclr.cc/paper_files/paper/2025/file/8c4de96b9169aa869cc102afe31055e8-Paper-Conference.pdf} {Bootstrapping language models with {DPO} implicit rewards}.
\newblock In \emph{International Conference on Learning Representations}, volume 2025.

\bibitem[{Chen et~al.(2021)Chen, Tworek, Jun, Yuan, de~Oliveira~Pinto, Kaplan, Edwards, Burda, Joseph, Brockman, Ray, Puri, Krueger, Petrov, Khlaaf, Sastry, Mishkin, Chan, Gray, Ryder, Pavlov, Power, Kaiser, Bavarian, Winter, Tillet, Such, Cummings, Plappert, Chantzis, Barnes, Herbert-Voss, Guss, Nichol, Paino, Tezak, Tang, Babuschkin, Balaji, Jain, Saunders, Hesse, Carr, Leike, Achiam, Misra, Morikawa, Radford, Knight, Brundage, Murati, Mayer, Welinder, McGrew, Amodei, McCandlish, Sutskever, and Zaremba}]{chen2021codexhumaneval}
Mark Chen, Jerry Tworek, Heewoo Jun, Qiming Yuan, Henrique~Ponde de~Oliveira~Pinto, Jared Kaplan, Harri Edwards, Yuri Burda, Nicholas Joseph, Greg Brockman, Alex Ray, Raul Puri, Gretchen Krueger, Michael Petrov, Heidy Khlaaf, Girish Sastry, Pamela Mishkin, Brooke Chan, Scott Gray, and 39 others. 2021.
\newblock \href {https://arxiv.org/abs/2107.03374} {Evaluating large language models trained on code}.
\newblock \emph{Preprint}, arXiv:2107.03374.

\bibitem[{Chen et~al.(2025{\natexlab{b}})Chen, He, Xi, Guo, Hong, Zhang, Li, Gui, Li, Zhang, and Huang}]{chen2025better}
Wenxiang Chen, Wei He, Zhiheng Xi, Honglin Guo, Boyang Hong, Jiazheng Zhang, Nijun Li, Tao Gui, Yun Li, Qi~Zhang, and Xuanjing Huang. 2025{\natexlab{b}}.
\newblock \href {https://doi.org/10.18653/v1/2025.findings-acl.747} {Better process supervision with bi-directional rewarding signals}.
\newblock In \emph{Findings of the Association for Computational Linguistics: ACL 2025}, pages 14471--14485, Vienna, Austria. Association for Computational Linguistics.

\bibitem[{Chen et~al.(2024)Chen, Deng, Yuan, Ji, and Gu}]{chen2024self}
Zixiang Chen, Yihe Deng, Huizhuo Yuan, Kaixuan Ji, and Quanquan Gu. 2024.
\newblock \href {https://proceedings.mlr.press/v235/chen24j.html} {Self-play fine-tuning converts weak language models to strong language models}.
\newblock In \emph{Proceedings of the 41st International Conference on Machine Learning}, volume 235 of \emph{Proceedings of Machine Learning Research}, pages 6621--6642. PMLR.

\bibitem[{Cheng et~al.(2025)Cheng, Xiong, Qiao, Li, Guo, Wang, Lv, and Wang}]{cheng2025pure}
Jie Cheng, Gang Xiong, Ruixi Qiao, Lijun Li, Chao Guo, Junle Wang, Yisheng Lv, and Fei-Yue Wang. 2025.
\newblock \href {https://openreview.net/forum?id=3Sxby0hH1q} {Stop summation: Min-form credit assignment is all process reward model needs for reasoning}.
\newblock In \emph{The Thirty-ninth Annual Conference on Neural Information Processing Systems}.

\bibitem[{Cui et~al.(2024)Cui, Yuan, Ding, Yao, He, Zhu, Ni, Xie, Xie, Lin, Liu, and Sun}]{cui2024ultrafeedback}
Ganqu Cui, Lifan Yuan, Ning Ding, Guanming Yao, Bingxiang He, Wei Zhu, Yuan Ni, Guotong Xie, Ruobing Xie, Yankai Lin, Zhiyuan Liu, and Maosong Sun. 2024.
\newblock \href {https://proceedings.mlr.press/v235/cui24f.html} {{ULTRAFEEDBACK}: Boosting language models with scaled {AI} feedback}.
\newblock In \emph{Proceedings of the 41st International Conference on Machine Learning}, volume 235 of \emph{Proceedings of Machine Learning Research}, pages 9722--9744. PMLR.

\bibitem[{Cui et~al.(2026)Cui, Yuan, Wang, Wang, Zhang, Chen, Li, He, Fan, Yu, Xu, Chen, Yuan, Chen, Zhang, Lv, Wang, Yao, Han, Peng, Cheng, Liu, Sun, Zhou, and Ding}]{cui2025processreinforcementimplicitrewards}
Ganqu Cui, Lifan Yuan, Zefan Wang, Hanbin Wang, Yuchen Zhang, Jiacheng Chen, Wendi Li, Bingxiang He, Yuchen Fan, Tianyu Yu, Qixin Xu, Weize Chen, Jiarui Yuan, Huayu Chen, Kaiyan Zhang, Xingtai Lv, Shuo Wang, Yuan Yao, Xu~Han, and 6 others. 2026.
\newblock \href {https://openreview.net/forum?id=9SkkifLopZ} {Process reinforcement through implicit rewards}.
\newblock \emph{Transactions on Machine Learning Research}.

\bibitem[{Deng and Mineiro(2024)}]{deng2024flow}
Yihe Deng and Paul Mineiro. 2024.
\newblock \href {https://arxiv.org/abs/2410.22304} {{Flow-DPO}: Improving {LLM} mathematical reasoning through online multi-agent learning}.
\newblock \emph{Preprint}, arXiv:2410.22304.

\bibitem[{Ding et~al.(2023)Ding, Chen, Xu, Qin, Hu, Liu, Sun, and Zhou}]{ding-etal-2023-enhancing}
Ning Ding, Yulin Chen, Bokai Xu, Yujia Qin, Shengding Hu, Zhiyuan Liu, Maosong Sun, and Bowen Zhou. 2023.
\newblock \href {https://doi.org/10.18653/v1/2023.emnlp-main.183} {Enhancing chat language models by scaling high-quality instructional conversations}.
\newblock In \emph{Proceedings of the 2023 Conference on Empirical Methods in Natural Language Processing}, pages 3029--3051, Singapore. Association for Computational Linguistics.

\bibitem[{Dubois et~al.(2024)Dubois, Liang, and Hashimoto}]{dubois2025lengthcontrolledalpacaevalsimpleway}
Yann Dubois, Percy Liang, and Tatsunori~B. Hashimoto. 2024.
\newblock \href {https://openreview.net/forum?id=CybBmzWBX0} {Length-controlled {AlpacaEval}: A simple way to debias automatic evaluators}.
\newblock In \emph{First Conference on Language Modeling}.

\bibitem[{Ethayarajh et~al.(2024)Ethayarajh, Xu, Muennighoff, Jurafsky, and Kiela}]{ethayarajh2024kto}
Kawin Ethayarajh, Winnie Xu, Niklas Muennighoff, Dan Jurafsky, and Douwe Kiela. 2024.
\newblock \href {https://proceedings.mlr.press/v235/ethayarajh24a.html} {Model alignment as prospect theoretic optimization}.
\newblock In \emph{Proceedings of the 41st International Conference on Machine Learning}, volume 235 of \emph{Proceedings of Machine Learning Research}, pages 12634--12651. PMLR.

\bibitem[{Gheshlaghi~Azar et~al.(2024)Gheshlaghi~Azar, Daniel~Guo, Piot, Munos, Rowland, Valko, and Calandriello}]{azar2024general}
Mohammad Gheshlaghi~Azar, Zhaohan Daniel~Guo, Bilal Piot, Remi Munos, Mark Rowland, Michal Valko, and Daniele Calandriello. 2024.
\newblock \href {https://proceedings.mlr.press/v238/gheshlaghi-azar24a.html} {A general theoretical paradigm to understand learning from human preferences}.
\newblock In \emph{Proceedings of The 27th International Conference on Artificial Intelligence and Statistics}, volume 238 of \emph{Proceedings of Machine Learning Research}, pages 4447--4455. PMLR.

\bibitem[{Grattafiori et~al.(2024)Grattafiori, Dubey, Jauhri, Pandey, Kadian, Al-Dahle, Letman, Mathur, Schelten, Vaughan, Yang, Fan, Goyal, Hartshorn, Yang, Mitra, Sravankumar, Korenev, Hinsvark, Rao, Zhang, Rodriguez, Gregerson, Spataru, Roziere, Biron, Tang, Chern, Caucheteux, Nayak, Bi, Marra, McConnell, Keller, Touret, Wu, Wong, Ferrer, Nikolaidis, Allonsius, Song, Pintz, Livshits, Wyatt, Esiobu, Choudhary, Mahajan, Garcia-Olano, Perino, Hupkes, Lakomkin, AlBadawy, Lobanova, Dinan, Smith, Radenovic, Guzmán, Zhang, Synnaeve, Lee, Anderson, Thattai, Nail, Mialon, Pang, Cucurell, Nguyen, Korevaar, Xu, Touvron, Zarov, Ibarra, Kloumann, Misra, Evtimov, Zhang, Copet, Lee, Geffert, Vranes, Park, Mahadeokar, Shah, van~der Linde, Billock, Hong, Lee, Fu, Chi, Huang, Liu, Wang, Yu, Bitton, Spisak, Park, Rocca, Johnstun, Saxe, Jia, Alwala, Prasad, Upasani, Plawiak, Li, Heafield, Stone, El-Arini, Iyer, Malik, Chiu, Bhalla, Lakhotia, Rantala-Yeary, van~der Maaten, Chen, Tan, Jenkins, Martin, Madaan, Malo, Blecher, Landzaat, de~Oliveira, Muzzi, Pasupuleti, Singh, Paluri, Kardas, Tsimpoukelli, Oldham, Rita, Pavlova, Kambadur, Lewis, Si, Singh, Hassan, Goyal, Torabi, Bashlykov, Bogoychev, Chatterji, Zhang, Duchenne, Çelebi, Alrassy, Zhang, Li, Vasic, Weng, Bhargava, Dubal, Krishnan, Koura, Xu, He, Dong, Srinivasan, Ganapathy, Calderer, Cabral, Stojnic, Raileanu, Maheswari, Girdhar, Patel, Sauvestre, Polidoro, Sumbaly, Taylor, Silva, Hou, Wang, Hosseini, Chennabasappa, Singh, Bell, Kim, Edunov, Nie, Narang, Raparthy, Shen, Wan, Bhosale, Zhang, Vandenhende, Batra, Whitman, Sootla, Collot, Gururangan, Borodinsky, Herman, Fowler, Sheasha, Georgiou, Scialom, Speckbacher, Mihaylov, Xiao, Karn, Goswami, Gupta, Ramanathan, Kerkez, Gonguet, Do, Vogeti, Albiero, Petrovic, Chu, Xiong, Fu, Meers, Martinet, Wang, Wang, Tan, Xia, Xie, Jia, Wang, Goldschlag, Gaur, Babaei, Wen, Song, Zhang, Li, Mao, Coudert, Yan, Chen, Papakipos, Singh, Srivastava, Jain, Kelsey, Shajnfeld, Gangidi, Victoria, Goldstand, Menon, Sharma, Boesenberg, Baevski, Feinstein, Kallet, Sangani, Teo, Yunus, Lupu, Alvarado, Caples, Gu, Ho, Poulton, Ryan, Ramchandani, Dong, Franco, Goyal, Saraf, Chowdhury, Gabriel, Bharambe, Eisenman, Yazdan, James, Maurer, Leonhardi, Huang, Loyd, Paola, Paranjape, Liu, Wu, Ni, Hancock, Wasti, Spence, Stojkovic, Gamido, Montalvo, Parker, Burton, Mejia, Liu, Wang, Kim, Zhou, Hu, Chu, Cai, Tindal, Feichtenhofer, Gao, Civin, Beaty, Kreymer, Li, Adkins, Xu, Testuggine, David, Parikh, Liskovich, Foss, Wang, Le, Holland, Dowling, Jamil, Montgomery, Presani, Hahn, Wood, Le, Brinkman, Arcaute, Dunbar, Smothers, Sun, Kreuk, Tian, Kokkinos, Ozgenel, Caggioni, Kanayet, Seide, Florez, Schwarz, Badeer, Swee, Halpern, Herman, Sizov, Guangyi, Zhang, Lakshminarayanan, Inan, Shojanazeri, Zou, Wang, Zha, Habeeb, Rudolph, Suk, Aspegren, Goldman, Zhan, Damlaj, Molybog, Tufanov, Leontiadis, Veliche, Gat, Weissman, Geboski, Kohli, Lam, Asher, Gaya, Marcus, Tang, Chan, Zhen, Reizenstein, Teboul, Zhong, Jin, Yang, Cummings, Carvill, Shepard, McPhie, Torres, Ginsburg, Wang, Wu, U, Saxena, Khandelwal, Zand, Matosich, Veeraraghavan, Michelena, Li, Jagadeesh, Huang, Chawla, Huang, Chen, Garg, A, Silva, Bell, Zhang, Guo, Yu, Moshkovich, Wehrstedt, Khabsa, Avalani, Bhatt, Mankus, Hasson, Lennie, Reso, Groshev, Naumov, Lathi, Keneally, Liu, Seltzer, Valko, Restrepo, Patel, Vyatskov, Samvelyan, Clark, Macey, Wang, Hermoso, Metanat, Rastegari, Bansal, Santhanam, Parks, White, Bawa, Singhal, Egebo, Usunier, Mehta, Laptev, Dong, Cheng, Chernoguz, Hart, Salpekar, Kalinli, Kent, Parekh, Saab, Balaji, Rittner, Bontrager, Roux, Dollar, Zvyagina, Ratanchandani, Yuvraj, Liang, Alao, Rodriguez, Ayub, Murthy, Nayani, Mitra, Parthasarathy, Li, Hogan, Battey, Wang, Howes, Rinott, Mehta, Siby, Bondu, Datta, Chugh, Hunt, Dhillon, Sidorov, Pan, Mahajan, Verma, Yamamoto, Ramaswamy, Lindsay, Lindsay, Feng, Lin, Zha, Patil, Shankar, Zhang, Zhang, Wang, Agarwal, Sajuyigbe, Chintala, Max, Chen, Kehoe, Satterfield, Govindaprasad, Gupta, Deng, Cho, Virk, Subramanian, Choudhury, Goldman, Remez, Glaser, Best, Koehler, Robinson, Li, Zhang, Matthews, Chou, Shaked, Vontimitta, Ajayi, Montanez, Mohan, Kumar, Mangla, Ionescu, Poenaru, Mihailescu, Ivanov, Li, Wang, Jiang, Bouaziz, Constable, Tang, Wu, Wang, Wu, Gao, Kleinman, Chen, Hu, Jia, Qi, Li, Zhang, Zhang, Adi, Nam, Yu, Wang, Zhao, Hao, Qian, Li, He, Rait, DeVito, Rosnbrick, Wen, Yang, Zhao, and Ma}]{grattafiori2024llama}
Aaron Grattafiori, Abhimanyu Dubey, Abhinav Jauhri, Abhinav Pandey, Abhishek Kadian, Ahmad Al-Dahle, Aiesha Letman, Akhil Mathur, Alan Schelten, Alex Vaughan, Amy Yang, Angela Fan, Anirudh Goyal, Anthony Hartshorn, Aobo Yang, Archi Mitra, Archie Sravankumar, Artem Korenev, Arthur Hinsvark, and 542 others. 2024.
\newblock \href {https://arxiv.org/abs/2407.21783} {The {Llama 3} herd of models}.
\newblock \emph{Preprint}, arXiv:2407.21783.

\bibitem[{Guo et~al.(2025)Guo, Yang, Zhang, Song, Wang, Zhu, Xu, Zhang, Ma, Bi, Zhang, Yu, Wu, Wu, Gou, Shao, Li, Gao, Liu, Xue, Wang, Wu, Feng, Lu, Zhao, Deng, Ruan, Dai, Chen, Ji, Li, Lin, Dai, Luo, Hao, Chen, Li, Zhang, Xu, Ding, Gao, Qu, Li, Guo, Li, Chen, Yuan, Tu, Qiu, Li, Cai, Ni, Liang, Chen, Dong, Hu, You, Gao, Guan, Huang, Yu, Wang, Zhang, Zhao, Wang, Zhang, Xu, Xia, Zhang, Zhang, Tang, Zhou, Li, Wang, Li, Tian, Huang, Zhang, Wang, Chen, Du, Ge, Zhang, Pan, Wang, Chen, Jin, Chen, Lu, Zhou, Chen, Ye, Wang, Yu, Zhou, Pan, Li, Zhou, Wu, Yun, Pei, Sun, Wang, Zeng, Liu, Liang, Gao, Yu, Zhang, Xiao, An, Liu, Wang, Chen, Nie, Cheng, Liu, Xie, Liu, Yang, Li, Su, Lin, Li, Jin, Shen, Chen, Sun, Wang, Song, Zhou, Wang, Shan, Li, Wang, Wei, Zhang, Xu, Li, Zhao, Sun, Wang, Yu, Zhang, Shi, Xiong, He, Piao, Wang, Tan, Ma, Liu, Guo, Ou, Wang, Gong, Zou, He, Xiong, Luo, You, Liu, Zhou, Zhu, Huang, Li, Zheng, Zhu, Ma, Tang, Zha, Yan, Ren, Ren, Sha, Fu, Xu, Xie, Zhang, Hao, Ma, Yan, Wu, Gu, Zhu, Liu, Li, Xie, Song, Pan, Huang, Xu, Zhang, and Zhang}]{guo2025deepseek}
Daya Guo, Dejian Yang, Haowei Zhang, Junxiao Song, Peiyi Wang, Qihao Zhu, Runxin Xu, Ruoyu Zhang, Shirong Ma, Xiao Bi, Xiaokang Zhang, Xingkai Yu, Yu~Wu, Z.~F. Wu, Zhibin Gou, Zhihong Shao, Zhuoshu Li, Ziyi Gao, Aixin Liu, and 175 others. 2025.
\newblock \href {https://doi.org/10.1038/s41586-025-09422-z} {{DeepSeek-R1} incentivizes reasoning in {LLMs} through reinforcement learning}.
\newblock \emph{Nature}, 645(8081):633--638.

\bibitem[{He et~al.(2024)He, Luo, Bai, Hu, Thai, Shen, Hu, Han, Huang, Zhang, Liu, Qi, Liu, and Sun}]{he2024olympiadbench}
Chaoqun He, Renjie Luo, Yuzhuo Bai, Shengding Hu, Zhen Thai, Junhao Shen, Jinyi Hu, Xu~Han, Yujie Huang, Yuxiang Zhang, Jie Liu, Lei Qi, Zhiyuan Liu, and Maosong Sun. 2024.
\newblock \href {https://doi.org/10.18653/v1/2024.acl-long.211} {{O}lympiad{B}ench: A challenging benchmark for promoting {AGI} with olympiad-level bilingual multimodal scientific problems}.
\newblock In \emph{Proceedings of the 62nd Annual Meeting of the Association for Computational Linguistics (Volume 1: Long Papers)}, pages 3828--3850, Bangkok, Thailand. Association for Computational Linguistics.

\bibitem[{Hendrycks et~al.(2021)Hendrycks, Burns, Kadavath, Arora, Basart, Tang, Song, and Steinhardt}]{hendrycksmath2021}
Dan Hendrycks, Collin Burns, Saurav Kadavath, Akul Arora, Steven Basart, Eric Tang, Dawn Song, and Jacob Steinhardt. 2021.
\newblock \href {https://datasets-benchmarks-proceedings.neurips.cc/paper_files/paper/2021/file/be83ab3ecd0db773eb2dc1b0a17836a1-Paper-round2.pdf} {Measuring mathematical problem solving with the {MATH} dataset}.
\newblock In \emph{Proceedings of the Neural Information Processing Systems Track on Datasets and Benchmarks}, volume~1.

\bibitem[{{Hugging Face}(2025)}]{openr1}
{Hugging Face}. 2025.
\newblock \href {https://github.com/huggingface/open-r1} {{Open R1}: A fully open reproduction of {DeepSeek-R1}}.

\bibitem[{Lambert et~al.(2025)Lambert, Pyatkin, Morrison, Miranda, Lin, Chandu, Dziri, Kumar, Zick, Choi, Smith, and Hajishirzi}]{lambert2024rewardbench}
Nathan Lambert, Valentina Pyatkin, Jacob Morrison, LJ~Miranda, Bill~Yuchen Lin, Khyathi Chandu, Nouha Dziri, Sachin Kumar, Tom Zick, Yejin Choi, Noah~A. Smith, and Hannaneh Hajishirzi. 2025.
\newblock \href {https://doi.org/10.18653/v1/2025.findings-naacl.96} {{R}eward{B}ench: Evaluating reward models for language modeling}.
\newblock In \emph{Findings of the Association for Computational Linguistics: NAACL 2025}, pages 1755--1797, Albuquerque, New Mexico. Association for Computational Linguistics.

\bibitem[{Lewkowycz et~al.(2022)Lewkowycz, Andreassen, Dohan, Dyer, Michalewski, Ramasesh, Slone, Anil, Schlag, Gutman-Solo, Wu, Neyshabur, Gur-Ari, and Misra}]{lewkowycz2022solving}
Aitor Lewkowycz, Anders Andreassen, David Dohan, Ethan Dyer, Henryk Michalewski, Vinay Ramasesh, Ambrose Slone, Cem Anil, Imanol Schlag, Theo Gutman-Solo, Yuhuai Wu, Behnam Neyshabur, Guy Gur-Ari, and Vedant Misra. 2022.
\newblock \href {https://doi.org/10.52202/068431-0278} {Solving quantitative reasoning problems with language models}.
\newblock In \emph{Advances in Neural Information Processing Systems}, volume~35, pages 3843--3857. Curran Associates, Inc.

\bibitem[{Li et~al.(2025)Li, Chiang, Frick, Dunlap, Wu, Zhu, Gonzalez, and Stoica}]{li2025arenahard}
Tianle Li, Wei-Lin Chiang, Evan Frick, Lisa Dunlap, Tianhao Wu, Banghua Zhu, Joseph~E. Gonzalez, and Ion Stoica. 2025.
\newblock \href {https://proceedings.mlr.press/v267/li25h.html} {From crowdsourced data to high-quality benchmarks: {Arena-Hard} and {Benchbuilder} pipeline}.
\newblock In \emph{Proceedings of the 42nd International Conference on Machine Learning}, volume 267 of \emph{Proceedings of Machine Learning Research}, pages 34209--34231. PMLR.

\bibitem[{Lightman et~al.(2024)Lightman, Kosaraju, Burda, Edwards, Baker, Lee, Leike, Schulman, Sutskever, and Cobbe}]{lightman2023let}
Hunter Lightman, Vineet Kosaraju, Yuri Burda, Harrison Edwards, Bowen Baker, Teddy Lee, Jan Leike, John Schulman, Ilya Sutskever, and Karl Cobbe. 2024.
\newblock \href {https://proceedings.iclr.cc/paper_files/paper/2024/file/aca97732e30bcf1303bc22ac3924fd16-Paper-Conference.pdf} {Let\textquotesingle s verify step by step}.
\newblock In \emph{International Conference on Learning Representations}, volume 2024.

\bibitem[{Liu et~al.(2024)Liu, Zhou, Liu, Bu, Yang, Zhong, and Ouyang}]{liu2024iterative}
Jie Liu, Zhanhui Zhou, Jiaheng Liu, Xingyuan Bu, Chao Yang, Han-Sen Zhong, and Wanli Ouyang. 2024.
\newblock \href {https://arxiv.org/abs/2406.11817} {Iterative length-regularized direct preference optimization: A case study on improving {7B} language models to {GPT-4} level}.
\newblock \emph{Preprint}, arXiv:2406.11817.

\bibitem[{Liu et~al.(2026)Liu, Zeng, Xiao, He, Liu, Wang, Yan, Shen, Zhang, Xu, and Liu}]{liu2025skywork}
Yuhao Liu, Liang Zeng, Yuzhen Xiao, Jujie He, Jiacai Liu, Chaojie Wang, Rui Yan, Wei Shen, Fuxiang Zhang, Jiacheng Xu, and Yang Liu. 2026.
\newblock \href {https://proceedings.iclr.cc/paper_files/paper/2026/file/d8836d0294651a867c2ca51012923997-Paper-Conference.pdf} {{Skywork-Reward-V2}: Scaling preference data curation via {Human-AI} synergy}.
\newblock In \emph{International Conference on Learning Representations}, volume 2026.

\bibitem[{Ma et~al.(2025)Ma, Liu, Jiang, Zhang, MA, and Chen}]{ma2025generalreasoneradvancingllmreasoning}
Xueguang Ma, Qian Liu, Dongfu Jiang, Ge~Zhang, Zejun MA, and Wenhu Chen. 2025.
\newblock \href {https://openreview.net/forum?id=pBFVoll8Xa} {{General-Reasoner}: Advancing {LLM} reasoning across all domains}.
\newblock In \emph{The Thirty-ninth Annual Conference on Neural Information Processing Systems}.

\bibitem[{Muennighoff et~al.(2025)Muennighoff, Yang, Shi, Li, Fei-Fei, Hajishirzi, Zettlemoyer, Liang, Cand{\`e}s, and Hashimoto}]{muennighoff2025s1simpletesttimescaling}
Niklas Muennighoff, Zitong Yang, Weijia Shi, Xiang~Lisa Li, Li~Fei-Fei, Hannaneh Hajishirzi, Luke Zettlemoyer, Percy Liang, Emmanuel Cand{\`e}s, and Tatsunori Hashimoto. 2025.
\newblock \href {https://doi.org/10.18653/v1/2025.emnlp-main.1025} {s1: Simple test-time scaling}.
\newblock In \emph{Proceedings of the 2025 Conference on Empirical Methods in Natural Language Processing}, pages 20275--20321, Suzhou, China. Association for Computational Linguistics.

\bibitem[{Ouyang et~al.(2022)Ouyang, Wu, Jiang, Almeida, Wainwright, Mishkin, Zhang, Agarwal, Slama, Ray, Schulman, Hilton, Kelton, Miller, Simens, Askell, Welinder, Christiano, Leike, and Lowe}]{ouyang2022traininglanguagemodelsfollow}
Long Ouyang, Jeffrey Wu, Xu~Jiang, Diogo Almeida, Carroll Wainwright, Pamela Mishkin, Chong Zhang, Sandhini Agarwal, Katarina Slama, Alex Ray, John Schulman, Jacob Hilton, Fraser Kelton, Luke Miller, Maddie Simens, Amanda Askell, Peter Welinder, Paul~F Christiano, Jan Leike, and Ryan Lowe. 2022.
\newblock \href {https://doi.org/10.52202/068431-2011} {Training language models to follow instructions with human feedback}.
\newblock In \emph{Advances in Neural Information Processing Systems}, volume~35, pages 27730--27744. Curran Associates, Inc.

\bibitem[{Rafailov et~al.(2023)Rafailov, Sharma, Mitchell, Manning, Ermon, and Finn}]{rafailov2023direct}
Rafael Rafailov, Archit Sharma, Eric Mitchell, Christopher~D Manning, Stefano Ermon, and Chelsea Finn. 2023.
\newblock \href {https://doi.org/10.52202/075280-2338} {Direct preference optimization: Your language model is secretly a reward model}.
\newblock In \emph{Advances in Neural Information Processing Systems}, volume~36, pages 53728--53741. Curran Associates, Inc.

\bibitem[{Rein et~al.(2024)Rein, Hou, Stickland, Petty, Pang, Dirani, Michael, and Bowman}]{rein2023gpqa}
David Rein, Betty~Li Hou, Asa~Cooper Stickland, Jackson Petty, Richard~Yuanzhe Pang, Julien Dirani, Julian Michael, and Samuel~R. Bowman. 2024.
\newblock \href {https://openreview.net/forum?id=Ti67584b98} {{GPQA}: A graduate-level {Google}-proof {Q\&A} benchmark}.
\newblock In \emph{First Conference on Language Modeling}.

\bibitem[{Schulman et~al.(2017)Schulman, Wolski, Dhariwal, Radford, and Klimov}]{ppo}
John Schulman, Filip Wolski, Prafulla Dhariwal, Alec Radford, and Oleg Klimov. 2017.
\newblock \href {https://arxiv.org/abs/1707.06347} {Proximal policy optimization algorithms}.
\newblock \emph{Preprint}, arXiv:1707.06347.

\bibitem[{Shao et~al.(2024)Shao, Wang, Zhu, Xu, Song, Bi, Zhang, Zhang, Li, Wu, and Guo}]{shao2024deepseekmathpushinglimitsmathematical}
Zhihong Shao, Peiyi Wang, Qihao Zhu, Runxin Xu, Junxiao Song, Xiao Bi, Haowei Zhang, Mingchuan Zhang, Y.~K. Li, Y.~Wu, and Daya Guo. 2024.
\newblock \href {https://arxiv.org/abs/2402.03300} {{DeepSeekMath}: Pushing the limits of mathematical reasoning in open language models}.
\newblock \emph{Preprint}, arXiv:2402.03300.

\bibitem[{Sheng et~al.(2025)Sheng, Zhang, Ye, Wu, Zhang, Zhang, Peng, Lin, and Wu}]{sheng2024hybridflow}
Guangming Sheng, Chi Zhang, Zilingfeng Ye, Xibin Wu, Wang Zhang, Ru~Zhang, Yanghua Peng, Haibin Lin, and Chuan Wu. 2025.
\newblock \href {https://doi.org/10.1145/3689031.3696075} {{HybridFlow}: A flexible and efficient {RLHF} framework}.
\newblock In \emph{Proceedings of the Twentieth European Conference on Computer Systems}, EuroSys \textquoteright{}25, pages 1279--1297. ACM.

\bibitem[{Singh et~al.(2024)Singh, Co-Reyes, Agarwal, Anand, Patil, Garcia, Liu, Harrison, Lee, Xu, Parisi, Kumar, Alemi, Rizkowsky, Nova, Adlam, Bohnet, Elsayed, Sedghi, Mordatch, Simpson, Gur, Snoek, Pennington, Hron, Kenealy, Swersky, Mahajan, Culp, Xiao, Bileschi, Constant, Novak, Liu, Warkentin, Bansal, Dyer, Neyshabur, Sohl-Dickstein, and Fiedel}]{singh2023beyond}
Avi Singh, John~D Co-Reyes, Rishabh Agarwal, Ankesh Anand, Piyush Patil, Xavier Garcia, Peter~J Liu, James Harrison, Jaehoon Lee, Kelvin Xu, Aaron~T Parisi, Abhishek Kumar, Alexander~A Alemi, Alex Rizkowsky, Azade Nova, Ben Adlam, Bernd Bohnet, Gamaleldin~Fathy Elsayed, Hanie Sedghi, and 21 others. 2024.
\newblock \href {https://openreview.net/forum?id=lNAyUngGFK} {Beyond human data: Scaling self-training for problem-solving with language models}.
\newblock \emph{Transactions on Machine Learning Research}.
\newblock Expert Certification.

\bibitem[{Tu et~al.(2025)Tu, Lin, Tian, Zhang, Li, Fu, Xu, He, Lan, Jiang, and Zhao}]{tu2025enhancing}
Songjun Tu, Jiahao Lin, Xiangyu Tian, Qichao Zhang, Linjing Li, Yuqian Fu, Nan Xu, Wei He, Xiangyuan Lan, Dongmei Jiang, and Dongbin Zhao. 2025.
\newblock \href {https://openreview.net/forum?id=OgWh4J7bkT} {Enhancing {LLM} reasoning with iterative {DPO}: A comprehensive empirical investigation}.
\newblock In \emph{Second Conference on Language Modeling}.

\bibitem[{Xie et~al.(2025)Xie, Gao, Ren, Luo, Hong, Dai, Zhou, Qiu, Wu, and Luo}]{xie2025logic}
Tian Xie, Zitian Gao, Qingnan Ren, Haoming Luo, Yuqian Hong, Bryan Dai, Joey Zhou, Kai Qiu, Zhirong Wu, and Chong Luo. 2025.
\newblock \href {https://arxiv.org/abs/2502.14768} {{Logic-RL}: Unleashing {LLM} reasoning with rule-based reinforcement learning}.
\newblock \emph{Preprint}, arXiv:2502.14768.

\bibitem[{Xiong et~al.(2024)Xiong, Dong, Ye, Wang, Zhong, Ji, Jiang, and Zhang}]{xiong2024iterative}
Wei Xiong, Hanze Dong, Chenlu Ye, Ziqi Wang, Han Zhong, Heng Ji, Nan Jiang, and Tong Zhang. 2024.
\newblock \href {https://proceedings.mlr.press/v235/xiong24a.html} {Iterative preference learning from human feedback: Bridging theory and practice for {RLHF} under {KL}-constraint}.
\newblock In \emph{Proceedings of the 41st International Conference on Machine Learning}, volume 235 of \emph{Proceedings of Machine Learning Research}, pages 54715--54754. PMLR.

\bibitem[{Xiong et~al.(2025)Xiong, Shi, Shen, Rosenberg, Qin, Calandriello, Khalman, Joshi, Piot, Saleh, Jin, Zhang, and Liu}]{xiong2024building}
Wei Xiong, Chengshuai Shi, Jiaming Shen, Aviv Rosenberg, Zhen Qin, Daniele Calandriello, Misha Khalman, Rishabh Joshi, Bilal Piot, Mohammad Saleh, Chi Jin, Tong Zhang, and Tianqi Liu. 2025.
\newblock \href {https://proceedings.iclr.cc/paper_files/paper/2025/file/40eff1670d6b08bb1bda48b0c5f30110-Paper-Conference.pdf} {Building math agents with multi-turn iterative preference learning}.
\newblock In \emph{International Conference on Learning Representations}, volume 2025.

\bibitem[{Yang et~al.(2025{\natexlab{a}})Yang, Li, Yang, Zhang, Hui, Zheng, Yu, Gao, Huang, Lv, Zheng, Liu, Zhou, Huang, Hu, Ge, Wei, Lin, Tang, Yang, Tu, Zhang, Yang, Yang, Zhou, Zhou, Lin, Dang, Bao, Yang, Yu, Deng, Li, Xue, Li, Zhang, Wang, Zhu, Men, Gao, Liu, Luo, Li, Tang, Yin, Ren, Wang, Zhang, Ren, Fan, Su, Zhang, Zhang, Wan, Liu, Wang, Cui, Zhang, Zhou, and Qiu}]{yang2025qwen3technicalreport}
An~Yang, Anfeng Li, Baosong Yang, Beichen Zhang, Binyuan Hui, Bo~Zheng, Bowen Yu, Chang Gao, Chengen Huang, Chenxu Lv, Chujie Zheng, Dayiheng Liu, Fan Zhou, Fei Huang, Feng Hu, Hao Ge, Haoran Wei, Huan Lin, Jialong Tang, and 41 others. 2025{\natexlab{a}}.
\newblock \href {https://arxiv.org/abs/2505.09388} {{Qwen3} technical report}.
\newblock \emph{Preprint}, arXiv:2505.09388.

\bibitem[{Yang et~al.(2024)Yang, Yang, Hui, Zheng, Yu, Zhou, Li, Li, Liu, Huang, Dong, Wei, Lin, Tang, Wang, Yang, Tu, Zhang, Ma, Yang, Xu, Zhou, Bai, He, Lin, Dang, Lu, Chen, Yang, Li, Xue, Ni, Zhang, Wang, Peng, Men, Gao, Lin, Wang, Bai, Tan, Zhu, Li, Liu, Ge, Deng, Zhou, Ren, Zhang, Wei, Ren, Liu, Fan, Yao, Zhang, Wan, Chu, Liu, Cui, Zhang, Guo, and Fan}]{yang2024qwen2}
An~Yang, Baosong Yang, Binyuan Hui, Bo~Zheng, Bowen Yu, Chang Zhou, Chengpeng Li, Chengyuan Li, Dayiheng Liu, Fei Huang, Guanting Dong, Haoran Wei, Huan Lin, Jialong Tang, Jialin Wang, Jian Yang, Jianhong Tu, Jianwei Zhang, Jianxin Ma, and 43 others. 2024.
\newblock \href {https://arxiv.org/abs/2407.10671} {{Qwen2} technical report}.
\newblock \emph{Preprint}, arXiv:2407.10671.

\bibitem[{Yang et~al.(2025{\natexlab{b}})Yang, Tong, Niu, Neubig, and Yue}]{yeo2025demystifying}
Shiming Yang, Yuxuan Tong, Xinyao Niu, Graham Neubig, and Xiang Yue. 2025{\natexlab{b}}.
\newblock \href {https://proceedings.mlr.press/v267/yang25ae.html} {Demystifying long chain-of-thought reasoning}.
\newblock In \emph{Proceedings of the 42nd International Conference on Machine Learning}, volume 267 of \emph{Proceedings of Machine Learning Research}, pages 71177--71209. PMLR.

\bibitem[{Ye et~al.(2025)Ye, Huang, Xiao, Chern, Xia, and Liu}]{ye2025limo}
Yixin Ye, Zhen Huang, Yang Xiao, Ethan Chern, Shijie Xia, and Pengfei Liu. 2025.
\newblock \href {https://openreview.net/forum?id=T2TZ0RY4Zk} {{LIMO}: Less is more for reasoning}.
\newblock In \emph{Second Conference on Language Modeling}.

\bibitem[{Yin et~al.(2026)Yin, Wei, Zhu, Chen, and Meng}]{yin-etal-2026-aligning}
Shangjian Yin, Zhepei Wei, Xinyu Zhu, Wei-Lin Chen, and Yu~Meng. 2026.
\newblock \href {https://doi.org/10.18653/v1/2026.acl-long.1595} {Aligning large language models via fully self-synthetic data}.
\newblock In \emph{Proceedings of the 64th Annual Meeting of the {A}ssociation for {C}omputational {L}inguistics (Volume 1: Long Papers)}, pages 34553--34568, San Diego, California, United States. Association for Computational Linguistics.

\bibitem[{Yu et~al.(2025)Yu, Yuan, Li, Xu, Wei, Wang, Qi, and Chen}]{yu2025longshortchainofthoughtmixturesupervised}
Bin Yu, Hang Yuan, Haotian Li, Xueyin Xu, Yuliang Wei, Bailing Wang, Weizhen Qi, and Kai Chen. 2025.
\newblock \href {https://arxiv.org/abs/2505.03469} {Long-short chain-of-thought mixture supervised fine-tuning eliciting efficient reasoning in large language models}.
\newblock \emph{Preprint}, arXiv:2505.03469.

\bibitem[{Zhang et~al.(2025)Zhang, Hosseini, Bansal, Kazemi, Kumar, and Agarwal}]{zhang2024generative}
Lunjun Zhang, Arian Hosseini, Hritik Bansal, Seyed~Mehran Kazemi, Aviral Kumar, and Rishabh Agarwal. 2025.
\newblock \href {https://proceedings.iclr.cc/paper_files/paper/2025/file/214308a2d5e3f83ef9ad2739e1cbc46d-Paper-Conference.pdf} {Generative verifiers: Reward modeling as next-token prediction}.
\newblock In \emph{International Conference on Learning Representations}, volume 2025.

\bibitem[{Zhang and Math-AI(2024)}]{aime24}
Yifan Zhang and Team Math-AI. 2024.
\newblock American invitational mathematics examination ({AIME}) 2024.

\bibitem[{Zhang and Math-AI(2025)}]{aime25}
Yifan Zhang and Team Math-AI. 2025.
\newblock American invitational mathematics examination ({AIME}) 2025.

\bibitem[{Zhang and Math-AI(2026)}]{aime26}
Yifan Zhang and Team Math-AI. 2026.
\newblock American invitational mathematics examination ({AIME}) 2026.

\end{thebibliography}
% \bibliographystyle{acl_natbib}
% !TEX root = acl.tex
% v2 appendix. The previous GRLO appendix is kept in appendix_grlo_v1.tex.
% Tables in tables/ are generated by scripts/v2/gen_tables.py.
\appendix

\section{Training Details}
\label{app:training}

Table~\ref{tab:hparams} lists the shared training configuration. Incomplete batches are dropped, so each epoch over a 5K pool consists of four batches of 1,024 prompts, drawn after reshuffling the pool. All runs in Table~\ref{tab:main} use it, except that runs on Web prompts allow prompts of up to 2,048 tokens because some WebInstruct questions are long. The reward-model ablation (Table~\ref{tab:reward}) uses the Chat-RM prompts and the shared configuration with Skywork-Reward-V2-Llama-3.2-3B or Skywork-Reward-V2-Qwen3-4B as the reward model. The data-scale runs (Table~\ref{tab:scale_main}) use 1K and 3K UltraFeedback prompts (984 unique prompts padded to 1,024 by repetition, and 2,946 prompts) for 15 epochs, i.e., 15 and 30 iterations; the 5K run is Chat-RM of Table~\ref{tab:main}. Each stage of the sequential runs (Table~\ref{tab:seq}) uses the shared configuration with 5K prompts for 15 epochs, starts from the final checkpoint of the previous stage, and initializes a new critic from that checkpoint; Math and Web stages use their verifiers and Chat stages the reward model. Math-RLVR and Math-RM use the same 5K MATH prompts in the same order and differ only in the reward. Training time is measured from the logs as the sum of per-iteration wall-clock times on one node with eight NVIDIA RTX PRO 6000 Blackwell GPUs (96\,GB each), excluding evaluation: 2.6 (Math-RLVR), 3.0 (Math-RM), 3.2 (Math+Chat-RM), 3.3 (Chat-RM), and 3.4 hours (Web-RLVR).

\begin{table}[!htbp]
    \centering
    \small
    \setlength{\tabcolsep}{4pt}
    \begin{tabular}{@{}ll@{}}
        \toprule
        Setting & Value \\
        \midrule
        Base model & Qwen3-4B-Base \\
        Framework & verl \\
        Algorithm & PPO with a value critic \\
        Advantage & GAE, $\gamma=1$, $\lambda=1$ \\
        Critic initialization & base model \\
        KL penalty / KL loss & none \\
        Entropy bonus & none \\
        Responses per prompt & 1 \\
        Rollout temperature / top-$p$ & 1.0 / 1.0 \\
        Prompt batch size & 1,024 \\
        PPO mini-batch size & 256 \\
        PPO epochs per batch & 1 \\
        Actor / critic learning rate & $10^{-6}$ / $10^{-5}$ \\
        Policy / value clip & 0.2 / 0.5 \\
        Max prompt / response length & 1,024 / 3,072 tokens \\
        Training epochs & 15 (60 iterations) \\
        \bottomrule
    \end{tabular}
    \caption{Shared training configuration.}
    \label{tab:hparams}
\end{table}

\section{Training Prompts}
\label{app:prompts}

\paragraph{Examples.}
Figure~\ref{fig:prompts} shows randomly drawn prompts from the three pools, lightly abridged. Math and Web prompts are questions with a single reference answer, whereas Chat prompts have no reference answer and span diverse domains, such as education, travel, and mathematics in Figure~\ref{fig:prompts}, as analyzed below.

\begin{figure*}[!htbp]
    \centering
    \small
    \fbox{\begin{minipage}[t]{0.465\linewidth}
    \raggedright
    \textbf{Math.} The digits 2, 3, 5 and 7 are arranged randomly to form a four-digit number. What is the probability that the number is odd? Express your answer as a common fraction. Let's think step by step and output the final answer within \texttt{\textbackslash boxed\{\}}.

    \smallskip
    \textbf{Web.} How much work (J) is done by friction when the mass of 150 kg is dragged for 10 m and if the coefficient of sliding friction is 0.3? Please reason step by step, and put your final answer within \texttt{\textbackslash boxed\{\}}.
    \end{minipage}}\hfill
    \fbox{\begin{minipage}[t]{0.465\linewidth}
    \raggedright
    \textbf{Chat.} Hi, I am a teacher who needs help crafting a science project for my secondary 1 students. Can I give you more information so you can help me better?

    \smallskip
    \textbf{Chat.} Could you recommend some affordable accommodations in Hong Kong?

    \smallskip
    \textbf{Chat.} Give step by step solved numerical example on Proper Orthogonal Decomposition.
    \end{minipage}}
    \caption{Example training prompts from the Math and Web pools (left) and the Chat pool (right).}
    \label{fig:prompts}
\end{figure*}

\paragraph{Domains of the Chat prompts.}
To characterize the Chat pool, we label every prompt with Gemini 3.8 Flash (temperature 0) by the domain of its subject matter rather than the task format, choosing among twelve domains (Table~\ref{tab:domains}). The pool is broad: the five largest domains are technology and computing (24.0\%), arts, literature, and entertainment (13.3\%), lifestyle, food, and travel (10.6\%), business and finance (10.2\%), and education and language (7.9\%), and the other seven domains together account for 34.0\%, none above 7.4\%. By source field, the prompts come from FLAN (34\%), ShareGPT (32\%), Evol-Instruct (16\%), UltraChat (15\%), and FalseQA (4\%). Mathematics and statistics account for only 4.4\% of the prompts, so Chat-RM sees few math problems during training, yet its MATH500 accuracy is close to that of Math-RLVR (79.2 vs.\ 79.8).

\begin{table}[!htbp]
    \centering
    \small
    \setlength{\tabcolsep}{4pt}
    \begin{tabular}{@{}lc@{}}
        \toprule
        Domain & Share \\
        \midrule
        Technology \& computing & 24.0 \\
        Arts, literature \& entertainment & 13.3 \\
        Lifestyle, food \& travel & 10.6 \\
        Business \& finance & 10.2 \\
        Education \& language & 7.9 \\
        Other domains (7) & 34.0 \\
        \bottomrule
    \end{tabular}
    \caption{Five largest subject-matter domains of the 5K Chat prompts (\%), labeled by Gemini 3.8 Flash.}
    \label{tab:domains}
\end{table}

\section{Evaluation Details}
\label{app:eval}

\paragraph{Verifiable benchmarks.}
MATH500 (500 problems), OlympiadBench (675), Minerva Math (272), AIME 2024, 2025, and 2026 (30 each), and GPQA-Diamond (198) are evaluated with the Qwen2.5-Math evaluation toolkit and its instruction prompt, which asks for step-by-step reasoning and a boxed final answer, with at most 8,192 new tokens. All of them use greedy decoding except AIME, for which we draw 32 samples per problem at temperature 0.6 and top-$p$ 0.95 and report the mean accuracy (avg@32) and the unbiased pass@$k$ estimator. HumanEval (164 problems) and MBPP (399 problems) use the chat-style protocol of \citet{cui2025processreinforcementimplicitrewards}, in which the problem is posed as a user instruction and a complete solution is required.

For HumanEval and MBPP, if a response does not parse as Python but parses after converting literal \texttt{\textbackslash n} and \texttt{\textbackslash t} sequences into newlines and tabs, we score the converted response. This can change scores; for example, for Qwen3-8B it recovers 239 MBPP and 10 HumanEval responses of the base model (MBPP 9.8 $\rightarrow$ 43.1, HumanEval 31.1 $\rightarrow$ 32.9) and a few responses of the 8B RLVR runs (at most +0.8 points).

Minerva scores additionally include numerical answers recovered from otherwise incorrect responses by extracting the numerical value and accepting a relative error of at most 1\%; strict scores are retained in the evaluation artifacts.

\paragraph{Open-ended benchmarks.}
AlpacaEval~2 (805 instructions) responses are sampled at temperature 0.6 and top-$p$ 0.95. The judge is Qwen2-72B-Instruct at temperature 0, the reference outputs are those of GPT-4 Turbo, and we report the length-controlled win rate. Arena-Hard v0.1 (500 prompts) responses use greedy decoding with at most 4,096 tokens and no system prompt. The judge is Qwen2-72B-Instruct at temperature 0, the baseline is GPT-4-0314, and each prompt is judged twice with the two responses swapped. Main scores are weighted pairwise win rates, with strong wins and losses weighted by three, averaged over 100 bootstrap replicates; intervals use the 5th and 95th percentiles. Pairs with an unparsed verdict are excluded. The separate style-control branch uses the official Bradley--Terry implementation with length and markdown features. For the second judge (\S\ref{sec:judge}), Gemini 3.8 Flash is queried at temperature 0 with low reasoning effort, using the same judge prompts, references, and scoring as for Qwen2-72B-Instruct.

\section{Additional Results}

% \subsection{Repeated training runs}
% \label{app:seeds}

% Chat-RM and Math+Chat-RM were each trained with two seeds (Table~\ref{tab:seeds}); Table~\ref{tab:main} reports the run marked with $^\ast$, and all other tables and figures use the same run unless stated otherwise. The two runs differ most on AE2 LC (up to 8.1 points), GPQA (up to 5.1), and MBPP (4.5), and by at most 2.5 points on MATH500 and Arena-Hard. Both Chat-RM runs stay well above the verifier runs on both open-ended benchmarks.

% \begin{table*}[!htbp]
%     \centering
%     \small
%     \setlength{\tabcolsep}{1.7pt}
%     \begin{tabular}{@{}lccccccccc@{}}
%         \toprule
%         Run & MATH500 & OlympiadBench & Minerva & AIME & GPQA & HumanEval & MBPP & AE2 LC & Arena-Hard \\
%         \midrule
%         \inputrows tables/app_seeds.tex
%         \bottomrule
%     \end{tabular}
%     \caption{Runs with two seeds and their mean $\pm$ standard deviation (\%). $^\ast$Run reported in Table~\ref{tab:main}.}
%     \label{tab:seeds}
% \end{table*}

\subsection{Repeated training runs}
\label{app:seeds}

Math+Chat-RM was trained with two random seeds (Table~\ref{tab:seeds}); Table~\ref{tab:main} reports the run marked with $^\ast$. The two runs differ by at most 2.1 points on every benchmark except GPQA (4.0 points), and both stay well above the verifier runs on the open-ended benchmarks.

\begin{table*}[!htbp]
    \centering
    \small
    \setlength{\tabcolsep}{1.7pt}
    \begin{tabular}{@{}lccccccccc@{}}
        \toprule
        Run & MATH500 & OlympiadBench & Minerva & AIME & GPQA & HumanEval & MBPP & AE2 LC & Arena-Hard \\
        \midrule
        Math+Chat-RM, run 1 & 80.6 & 42.2 & 48.2 & 14.4 & 45.0 & 79.3 & 56.6 & 52.5 & 77.0 \\
        Math+Chat-RM, run 2$^\ast$ & 78.6 & 44.3 & 46.3 & 12.9 & 49.0 & 81.1 & 56.6 & 54.0 & 76.6 \\
        Math+Chat-RM, mean & 79.6\sd{1.4} & 43.3\sd{1.5} & 47.3\sd{1.3} & 13.7\sd{1.1} & 47.0\sd{2.8} & 80.2\sd{1.3} & 56.6\sd{0.0} & 53.3\sd{1.1} & 76.8\sd{0.3} \\
        \bottomrule
    \end{tabular}
    \caption{Math+Chat-RM with two seeds and their mean $\pm$ standard deviation (\%). $^\ast$Run reported in Table~\ref{tab:main}.}
    \label{tab:seeds}
\end{table*}

\subsection{Results across training epochs}
\label{app:epochs}

Table~\ref{tab:epochs} reports checkpoints after 1, 5, 10, and 15 epochs; these are the data behind Figure~\ref{fig:dynamics}, and epoch 15 matches Table~\ref{tab:main} except for AIME, which is reported here as avg@8 over AIME 2024 and 2025. Epoch-1 checkpoints come from separate one-epoch runs with the same configuration. After one epoch, scores vary widely across runs (HumanEval 7.3 for Math-RLVR, 20.1 for Chat-RM, 27.4 for Web-RLVR, and 73.2 for Math-RM); these early numbers mostly reflect how quickly each run learns the response format. From epoch 5 on, code and math scores are similar across runs, while AE2 LC grows with the share of open-ended prompts under the reward model; the exception is Web-RLVR+Chat-RM, whose MATH500, HumanEval, and AE2 LC dip at epoch 5 before recovering.

\begin{table*}[t]
    \centering
    \small
    \setlength{\tabcolsep}{2.5pt}
    \begin{tabular}{@{}llccccccc@{}}
        \toprule
        Run & Epoch & MATH500 & AIME & GPQA & HumanEval & MBPP & AE2 WR & AE2 LC \\
        \midrule
        \inputrows tables/app_epochs.tex 
        \bottomrule
    \end{tabular}
    \caption{Results across training epochs (\%). AE2 WR is the raw AlpacaEval~2 win rate. AIME is avg@8 over the 2024 and 2025 exams. ``--'': not evaluated.}
    \label{tab:epochs}
\end{table*}

\subsection{Pass@$k$}
\label{app:passk}

Table~\ref{tab:passk} complements Figure~\ref{fig:passk} with all configurations. Values are pooled over AIME 2024, 2025, and 2026, so pass@1 equals the AIME column of Table~\ref{tab:main}.

\begin{table}[!htbp]
    \centering
    \small
    \setlength{\tabcolsep}{2.1pt}
    \begin{tabular}{@{}lcccccc@{}}
        \toprule
        Model & $k{=}1$ & 2 & 4 & 8 & 16 & 32 \\
        \midrule
        \inputrows tables/app_passk.tex 
        \bottomrule
    \end{tabular}
    \caption{Unbiased pass@$k$ (\%) pooled over AIME 2024, 2025, and 2026 (90 problems, 32 samples per problem).}
    \label{tab:passk}
\end{table}

\begin{figure*}[!htbp]
    \centering
    \includegraphics[width=0.82\linewidth]{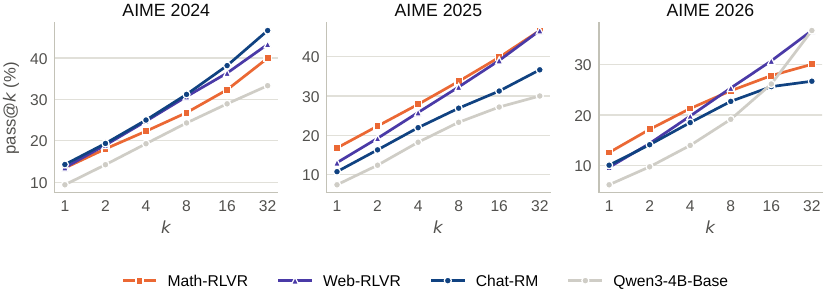}
    \caption{Unbiased pass@$k$ on each AIME exam (30 problems, 32 samples per problem).}
    \label{fig:passk_exam}
\end{figure*}

Figure~\ref{fig:passk_exam} shows the curves for each exam separately. At small $k$, all three runs improve on the base model on every exam, but none is consistently ahead: Chat-RM leads on AIME 2024 and Math-RLVR on 2025 and 2026. At $k{=}32$ the gains depend on the exam; on AIME 2026 the base model (36.7) ties Web-RLVR and overtakes Math-RLVR (30.0) and Chat-RM (26.7). With 30 problems per exam, a single problem is worth 3.3 points.

\subsection{MATH500 error decomposition}
\label{app:errors}

For every incorrect MATH500 response, Qwen3-4B-Instruct-2507 judges whether the error lies in the answer format or in the reasoning. Because this judge labels almost every response without an extractable answer as a format error, including degenerate ones, we apply two rules afterwards. Degenerate outputs (a 2--30 character span repeated at least nine times, leaked chat-template tokens, or a restated problem) are counted as reasoning errors, and a format error is kept only if the extracted answer matches the reference after removing units or the reference appears in the last 400 characters of the response. The second rule can accept coincidental matches of very short answers, so we inspected by hand every format error of the five models in Table~\ref{tab:errors_main} and reassigned three truncated, unfinished responses (two of the base model and one of Math-RLVR) to reasoning errors. Format errors are rare for every model, and the gains of RL come from fewer reasoning errors.

\begin{table}[!htbp]
    \centering
    \small
    \setlength{\tabcolsep}{3.5pt}
    \begin{tabular}{@{}lccc@{}}
        \toprule
        Model & Accuracy & Format & Reasoning \\
        \midrule
        \inputrows tables/app_errors.tex 
        \bottomrule
    \end{tabular}
    \caption{MATH500 accuracy and error types as percentages of the 500 problems.}
    \label{tab:errors}
\end{table}

\begin{table}[!htbp]
    \centering
    \small
    \setlength{\tabcolsep}{1.9pt}
    \begin{tabular}{@{}lccccc@{}}
        \toprule
        Model & Mean & Median & Capped & WR & WR$_{\text{sc}}$ \\
        \midrule
        \inputrows tables/app_length.tex 
        \bottomrule
    \end{tabular}
    \caption{Arena-Hard response length in tokens, the percentage of responses reaching the 4,096-token cap, and the win rate without and with style control (sc).}
    \label{tab:length}
\end{table}

\begin{table*}[!htbp]
    \centering
    \small
    \setlength{\tabcolsep}{1.6pt}
    \renewcommand{\arraystretch}{1.1}
    \begin{tabular*}{\linewidth}{@{\extracolsep{\fill}}l ccccccccc@{}}
\toprule
& \multicolumn{7}{c}{Verifiable domains} & \multicolumn{2}{c}{Open-ended domains} \\
\cmidrule(lr){2-8}\cmidrule(l){9-10}
Model & MATH500 & OlympiadBench & Minerva & AIME & GPQA & HumanEval & MBPP & AE2 LC & Arena-Hard \\
\midrule
Llama-3.2-3B-Instruct & 43.6 & 15.0 & 20.6 & 2.8 & 10.1 & 54.9 & 51.4 & 20.8 & 26.0 \\
\midrule
Math-RLVR & 46.8 & 14.1 & 25.4 & 3.5 & 19.2 & \textbf{56.7} & \textbf{56.6} & 23.3 & 27.2 \\
Web-RLVR & 42.6 & \textbf{16.7} & 25.0 & 2.2 & \textbf{20.2} & 54.3 & 54.4 & 23.0 & 25.0 \\
Math-RM & \textbf{47.8} & 16.4 & \textbf{28.3} & \textbf{3.7} & 15.2 & 56.1 & 54.4 & 29.9 & 31.4 \\
\midrule
Chat-RM & 47.2 & 16.4 & 25.0 & 2.5 & 19.2 & \textbf{56.7} & 51.6 & \textbf{35.1} & \textbf{45.9} \\
\bottomrule
\end{tabular*}

    \caption{RL from Llama-3.2-3B-Instruct (\%), with the recipe and evaluation of Table~\ref{tab:main}. Bold marks the best score in each column.}
    \label{tab:llama}
\end{table*}

\subsection{Response length and style control}
\label{app:length}

A natural concern is that reward-model RL raises open-ended scores simply by producing longer responses. Figure~\ref{fig:length} shows that RL lengthens responses only on open-ended prompts. On MATH500, every RL run stays within 433--557 tokens, close to the base model (466). On AlpacaEval, the verifier runs write shorter responses than the base model (434--440 vs.\ 668 tokens), whereas under the reward model the length grows with how open-ended the training prompts are, from 400 tokens (Math-RM) to 528 (Web-RM), 665 (Math+Chat-RM), and 729 (Chat-RM). Length alone does not explain the gains, however, since AE2 LC already controls for length.

Table~\ref{tab:length} reports the length of Arena-Hard responses and the win rate with and without style control, which regresses out response length and markdown features. Runs with open-ended prompts write longer answers, and style control lowers their win rates by 2.4--2.9 points but preserves the ordering. The base model has no style-control score because its responses lack the metadata the official implementation requires.

\subsection{Llama-3.2-3B-Instruct}
\label{app:llama}

To test another model family under the present recipe, we train Math-RLVR, Web-RLVR, Math-RM, and Chat-RM from Llama-3.2-3B-Instruct~\citep{grattafiori2024llama} with the same prompts, rewards, and hyperparameters as on Qwen3-4B-Base (Table~\ref{tab:llama}). We start from the instruction-tuned model because Llama-3.2-3B-Base does not follow instructions reliably. The results follow the Qwen3 pattern. Verifiable gains are smaller for this model but similar across the four runs (MATH500 42.6--47.8 vs.\ 43.6 for the starting model; Minerva 25.0--28.3 vs.\ 20.6), and Chat-RM is on par with the verifier runs except on MBPP (51.6 vs.\ 54.4--56.6). Open-ended quality instead depends on the reward and the prompts: with a verifier it stays near the starting model (AE2 LC 23.0--23.3 vs.\ 20.8; Arena-Hard 25.0--27.2 vs.\ 26.0), whereas with the reward model it rises to 29.9 and 31.4 on Math prompts and to 35.1 and 45.9 on Chat prompts.

% \subsection{Mixed-reward runs}
% \label{app:mix}

% Besides Web-RLVR+Chat-RM in Table~\ref{tab:main}, we trained seven other mixed-reward runs (Web prompts scored by the verifier and Chat prompts by the reward model, half of each) with 1K to 5K prompts and the recipe of Table~\ref{tab:hparams}. Two of the eight runs diverged: their entropy rose to 3.4--4.6 nats per token within the first iterations, gradient norms reached 8--13 (about 0.4 for Chat-RM), and the outputs became repetitive, mixed-language, or leaked chat-template tokens. A third run degraded at epoch 10 and recovered by epoch 15, and Web-RLVR+Chat-RM temporarily answered almost half of the MATH500 problems in Chinese at epoch 5. Runs with identical settings both diverged and trained stably, so the divergence is not explained by a configuration difference.

\end{document}